\documentclass[preprint,12pt,authoryear]{elsarticle}
\usepackage[utf8]{inputenc}

\usepackage{amssymb}
\usepackage{amsmath}
\usepackage{tabularx}
\usepackage{lipsum}
\usepackage{multirow}
\usepackage{graphicx}
\usepackage{svg}
\usepackage[table]{xcolor}
\usepackage{dblfloatfix}  
\usepackage{float}
\usepackage{natbib}
\usepackage{booktabs}
\usepackage{xurl}
\usepackage{pdflscape}
\usepackage[hidelinks]{hyperref}
\usepackage{siunitx}
\usepackage[utf8]{inputenc}
\usepackage{lineno}

\newcommand{\EG}{\textit{e.g.,~}} 
\newcommand{\IE}{\textit{i.e.,~}}
\newcommand{\lidar}{LiDAR}
\newcommand{\Lon}{leaf-on}
\newcommand{\ULon}{Leaf-on}
\newcommand{\Loff}{leaf-off}
\newcommand{\ULoff}{Leaf-off}
\newcommand{\MaMethode}{\texttt{FLORA}}

\journal{Remote Sensing of Environment}

\begin{document}

\begin{frontmatter}



\title{\MaMethode: A deep learning approach to predict forest attributes from heterogeneous LiDAR data} 

 \author[label1,label2]{Emilie Vautier\corref{cor1}}
 \affiliation[label1]{organization={Univ Gustave Eiffel, Géodata Paris, IGN, LASTIG},
          postcode={F-77454 Marne-la-Vallée},
             country={France}}
 \affiliation[label2]{organization={Univ Gustave Eiffel, Géodata Paris, IGN, LIF},
             postcode={F-54000, Nancy},
             country={France}}
 \affiliation[label3]{organization={Université de Lorraine, Géodata Paris, IGN, LIF},
          postcode={F-54000 Nancy},
             country={France}}
\author[label1]{Clément Mallet} 
\author[label2,label3]{Cédric Vega}
\cortext[cor1]{Corresponding author: Emilie Vautier. Email: emilie.vautier1@gmail.com}
\begin{abstract}

Forest attributes are essential for national-scale resource monitoring. Among auxiliary data supporting National Forest Inventory (NFI) estimates, airborne LiDAR metrics are among the most strongly correlated with forest attributes. However, producing operational wall-to-wall predictions remains challenging when LiDAR data are acquired under heterogeneous conditions. As national LiDAR programs expand across Europe, variability in flight constraints, sensors, seasons, and scan angles becomes a major obstacle to consistent forest attribute mapping. Existing models are often local and struggle to generalize beyond their calibration domain, especially at national scale over structurally diverse forests.

We present \MaMethode\ (Forest LiDAR Octree Regression with Auxiliary Data), a deep learning framework predicting six forest attributes: dominant height, total volume, deciduous and coniferous volumes, basal area, and stem density, from heterogeneous LiDAR point clouds. \MaMethode\ combines an octree-based high-resolution backbone with ecological and spatiotemporal auxiliary variables through a late-fusion gating mechanism, enabling partial interpretability of contextual data contributions for each forest attribute prediction. Models are trained and evaluated using 32,052 National Forest Inventory plots across mainland France, using data from the French LiDAR HD program covering diverse forest types and acquisition conditions.

Experiments show that a single model trained jointly on leaf-on and leaf-off acquisitions outperforms season-specific models and improves cross-season robustness. Auxiliary variables provide modest overall gains but reveal attribute-dependent contributions, particularly for species-specific volume prediction. The model achieves strong performance for dominant height  (rRMSE $\approx12.3\%$, R² $\approx 0.88$) and competitive results for total volume (rRMSE $\approx 39\%$, R² $\approx 0.74$). These results establish \MaMethode\ as a robust baseline for large-scale forest attribute estimation from national LiDAR programs.

\end{abstract}


\begin{highlights}
\item \lidar\ acquisition variability affects forest attributes predictions.
\item Nationwide French \lidar\ dataset enable large-scale heterogeneous analysis.
\item Deep learning exploits \lidar\ point clouds while preserving forest structure for forest attributes estimation.
\item Season-specific trained models fail across different phenology.
\item Multi-season training improves robustness and national-scale transferability.
\end{highlights}


\end{frontmatter}


\section{Introduction}
\label{sec:intro}


National Forest Inventories (NFIs) are designed to estimate and sometimes to monitor the change in forests characteristics at National scale with a given level of reliability. In order to downscale NFI estimates without increasing the cost of field surveys while preserving the reliability of the estimates, methods have been developed to combine NFI data with auxiliary data. Those so-called Multisource Forest Inventories (MSNFIs) usually rely on remote sensing data selected for their correlation with the forest attributes of interest, their low cost, and their availability at the national scale, together with their regular long-term renewal \citep{tomppo_combining_2008}. In terms of correlation, 3D data derived from airborne laser scanning, stereophotogrammetry, and, to a lesser extent, radar interferometry are the most suitable sources of auxiliary information for structural forest attributes such as volume and biomass \citep{borsah_lidar-based_2023}. With the development of national-scale \lidar\ databases, interest in their use within MSNFIs has been renewed. That said, data acquisition and processing are not standardized, making their integration into MSNFIs a challenge.

\lidar\ data processing methods are established for point clouds acquired locally, and often with similar acquisition conditions. Nevertheless, a significant challenge arises with national acquisitions, which often rely on  point clouds acquired under varying acquisition conditions \citep{petras_point_2023}.
Early studies have pointed out that \lidar\ acquisition parameters such as flying altitude, scan angle, pulse density, and sensor characteristics influence the estimation of forest structural attributes.
\citet{naesset_effects_2009} showed, from four Airborne Laser Scanning (ALS) acquisitions over 40 field plots in mature conifer forest, that instrument type, flying altitude, and pulse repetition frequency all affect ALS-derived height and density metrics, with systematic differences between acquisitions of up to 2.5\% for mean tree height and 10.7\% for timber volume, despite only minor differences in overall model precision. \citet{gopalakrishnan_prediction_2015} further demonstrated, using linear regression on $\sim$1,800 field plots across 76 heterogeneous ALS projects in the Southeastern US, that combining heterogeneous \lidar\ datasets over large areas introduces additional variability, with factors such as point density, scan angle, and vegetation heterogeneity influencing model accuracy, though canopy height could still be predicted with a root mean squared error (RMSE) of 3.0$\:$m (R$^2$ = 0.74), reduced to 2.4 m$\:$in structurally homogeneous stands.
\citet{qin_simulating_2017} specifically highlighted the scan angle as a critical parameter for foliage profile retrieval, identifying an optimal angle of 20$^{\circ}$ regardless of flying altitude or pulse density, while showing that combining multiple scan angles can further improve estimation accuracy.

Differences between \Lon\ and \Loff\ acquisition conditions have been shown to substantially affect canopy structure representation and the estimation of forest attributes. In \textit{deciduous} stands in particular, the absence of foliage leads to a greater penetration of the signal through the canopy, resulting in a higher number of single and last returns originating from the ground and a shift of the return distribution toward lower heights under \Loff\ conditions \citep{orka_effects_2010}. These structural differences have direct consequences for model transferability: applying a k-nearest neighbors (kNN) algorithm, a non-parametric method that predicts values based on the most similar observations, to predict volume-related values by analyzing the volumes of the most similar data points, trained under \Lon\ conditions to \Loff\ data (and vice versa) introduces substantial bias and should therefore be avoided \citep{villikka_suitability_2012}. 
This degradation in performance has been quantified in boreal forest contexts using Random Forest models across two inventory areas in Finland, where cross-using \Lon\ and \Loff\ models increased relative Root Mean Square Error (rRMSE)  and led to both over and underestimation, particularly in \textit{deciduous}-dominated plots. In these cases, error nearly doubled (from $\sim$33\% to $\sim$56\%) and bias reached up to 29\%, although calibration via an empirical ratio estimator was shown to partially mitigate these errors \citep{maltamo_transferability_2025}. Consistent with these findings, further evidence shows that cross-application of random forest models and acquisition conditions results in large increases in error for both forest types, with average absolute increases of rRMSE up to 16.8\% for coniferous stands and 25.8\% for deciduous stands, reinforcing that such mismatches should be avoided \citep{white_evaluating_2015}. These results highlight the critical importance of accounting for phenological conditions when developing and applying \lidar-based forest attribute models.

The reliability of estimates also vary with the forest type (even at a coarse level such as \textit{coniferous}, \textit{deciduous}, and \textit{mixed forests}). For \textit{deciduous} trees, leaf conditions influence the distribution of points, with skewed distributions toward the upper crown in \Lon\ conditions with limited penetration to the ground, to shifted point distribution toward lower canopy levels with increased ground sampling in \Loff\ conditions \citep{davison_effect_2020}. Pure \textit{coniferous} stands tend to be minimally impacted  by acquisition conditions \citep{wasser_influence_2013}, albeit the residual presence of broadleaf trees, lower vegetation and partial winter needle thinning \citep{davison_effect_2020} still introduces variability. One can conclude that most recent findings and models are heavily area- and acquisition-dependent \citep{fareed_interdisciplinary_2026}.

Taken together, these acquisition-related and phenological sources of variability make it particularly challenging to develop forest variable estimation models that generalize beyond their local or regional scales, and the conditions under which they were trained. This motivates the development of more robust, large-scale approaches.\\

Estimating forest attributes at national scale with a single, generic model capable of handling a large diversity of forest types and \lidar\ acquisition settings remains, to the best of our knowledge, an open problem as to date, the authors are not aware of such investigations. In this paper, we address this challenge by leveraging the French national \lidar\ HD program. Such acquisition program provides a uniquely large and heterogeneous dataset of 3D point clouds, with minimum density of 10$\:$pts/m$^2$ and acquisitions distributed across both \Loff\ and \Lon\ conditions. French forests are dominated at 68\% by \textit{deciduous} stands, a forest type whose structural signal is strongly modulated by phenology: such dataset captures a variability rarely encountered in previous studies. This heterogeneity is further compounded by the five-year acquisition window, which spans multiple campaigns conducted under different seasonal and instrumental conditions, making the dataset not only large but intrinsically variable.\\

Harmonization remains, to our knowledge, largely unexplored in this context, although several strategies have been proposed in the literature to address acquisition heterogeneity.
One approach consists in developing separate estimation models for each acquisition campaign or season, as implemented by \citet{nilsson_nationwide_2017} within the Swedish NFI framework:  block-specific (usually 25$\:$km by 50$\:$km in size) linear regression models were fitted using the nearest NFI plots to account for the variability introduced by different ALS sensors and acquisition settings across the country. Alternatively, linear mixed-effects models have been proposed to account for both forest type and acquisition variability, as demonstrated by \citet{hauglin_large_2021} in the context of the Norwegian forest resource map, where data from 367 ALS projects acquired over ten years were combined with NFI plots through stratification by species and maturity class. Species-specific models yielded rRMSE of 35\%, 34\%, 31\% and 12\% for volume, aboveground biomass, basal area and Lorey's height, respectively, outperforming general unstratified models by 2–7 percentage points.

However, explicitly stratifying models by acquisition condition and forest type remains challenging at national scale, where the compounding diversity of forest composition, climate, and topography, combined with the difficulty of aggregating small areas due to error propagation \citep{walshe_investigating_2021,saarela_hierarchical_2016,breidenbach_quantifying_2014} limits the practical applicability of such approaches. The question of spatial transferability further illustrates these limitations: \citet{soininen_transferability_2025} proposed a Random Forest approach combining ALS and NFI data across Finland, yet reported a systematic increase in RMSE with geographic distance from the training set, highlighting the limited generalization capacity of such models under spatial heterogeneity. Moreover, phenology represents only one dimension of the problem: flight altitude, sensor type, and point density each influence point cloud properties in ways that interact with forest type and structure. This renders exhaustive stratification both impractical and potentially inadequate. A more robust solution requires a form of \textit{harmonization}, whether applied at the input data level (making 3D point clouds similar in terms of geometry) or at the prediction level (estimating consistent forest variable whatever the above-mentioned fluctuations).

One straightforward approach would consist in homogenizing datasets by converting all 3D point clouds into a specific targeted condition (\EG \Lon). However, beyond phenology, many other acquisition characteristics like flight altitude, sensor type, density as well as their interaction with each forest type and structure have been shown to affect point cloud properties \citep{jakubowski_tradeoffs_2013,larue_evaluating_2022}. Explicitly normalizing for every known effect is a complex reconstruction task, and likely impractical for large-scale operational use on heterogeneous datasets \citep{ruetschi_countrywide_2021}. It is also based on the assumption that it is possible to retrieve the correct geometry of forests. However, the literature mainly focuses on point cloud completion and has only been addressed at the individual tree level or for opaque surfaces \citep{bornand_completing_2024,luo_inceptionformer_2026}.

Instead, another approach could focus on predicting similar outputs. It assumes a statistical model can capture such heterogeneity and can automatically learn to adjust its prediction to the acquisition and seasonal context. This hypothesis is supported by \citet{white_evaluating_2015}. They showed, using Random Forest models trained on 787 ground plots in a lodgepole pine-dominated forest in Alberta, Canada, that pooling \Lon\ and \Loff\ data into a single model results in only marginal performance loss (<2\% rRMSE for coniferous, near-zero for deciduous stands). This suggests that a single model can implicitly accommodate both acquisition conditions without explicit stratification.
At larger scales, \citet{hauglin_large_2021} further demonstrated that such an approach remains feasible even under strong acquisition heterogeneity, successfully combining data from 367 ALS projects acquired over a decade across 17$\:$Mha of Norwegian forest within a single modeling framework.\\

Traditional machine learning approaches rely on pairing carefully designed statistical features derived from point clouds with ground-truth measurements from forest inventory plots to perform predictions. Regression models for forest attributes (volume, biomass, basal area, etc.), typically based on Random forest and kNN algorithms, often struggle to generalize when applied over heterogeneous conditions. Even if methods like kNN have the advantage of providing confidence intervals using empirical quantiles of the neighbors or bootstrapping \citep{sagar_forest_2025}, they are still limited by their ability to learn robust transferable representation \citep{villikka_suitability_2012,parra_fiabilite_2025}. Recent advances in deep learning (DL) offer new perspectives to overcome these limitations. \citet{oehmcke_deep_2023} highlight that DL regression of forest attributes can produce more accurate results compared to state-of the art machine learning methods.
By operating directly on raw point clouds, DL models preserve the full three-dimensional structure of the data and automatically learn relevant spatial features, bypassing the need for manual feature selection, and fluctuation in the local geometry of point clouds. 

Several DL architectures have been proposed to process 3D point clouds, broadly falling into two categories: point-based and voxel-based approaches. Point-based approaches rely on sampling or grouping strategies to capture local context \citep{wu_point_2024,geist_ez-sp_2025}, a strategy that is not well suited to the structural irregularity of forest canopies, where vegetation density varies continuously and no clear object boundaries exist. Voxel-based discretizations, in contrast, preserve structural information while 
maintaining computational efficiency, and are better adapted to the sparse and heterogeneous nature of forest point clouds.\\
\citet{seely_modelling_2023} evaluated two DL architectures for forest attribute regression in a temperate mixed forest: the point-based Dynamic Graph Convolutional Neural Network (DGCNN) and the voxel-based Octree Convolutional Neural Network (OCNN). Their results show that while both DL methods outperformed Random Forests, the voxel-based OCNN achieved the lowest prediction errors and reduced mean absolute percentage error.
\citet{oehmcke_deep_2024}, in their study on airborne \lidar\ biomass estimation over Danish national forests, adapted and compared three distinct DL architectures for regression: the point-based PointNet, the point-based Kernel Point Convolution (KPConv), and the voxel-based Minkowski convolutional neural network. Among these, the voxel-based Minkowski CNN consistently outperformed both point-based alternatives across all forest attributes and evaluation metrics.
Both studies highlight that (i) voxel-based architectures exhibit greater robustness to variations in forest type, (ii) reliable predictions are conceivable by training a DL model with multi-annual NFI and \lidar\ data that may not be fully temporally aligned.\\

In this paper, we propose a generic deep-learning  solution, dubbed \MaMethode\ (Forest \lidar\ Octree Regression with Auxiliary data), for estimating forest attributes at a national-level, that can handle a large diversity of both forest and \lidar\ data conditions usually encountered at national scales. \MaMethode\ is a voxel-based approach designed to estimate forest structural attributes from heterogeneous \lidar\ data, validated over France where large-scale NFI and \lidar\ data are jointly available. The proposed data does not have counterpart of this size so far \citep{wegen_survey_2025}. Our model predicts six attributes commonly derived from \lidar\ data \citep{coops_modelling_2021}. We focused here on the following six : dominant height, total volume, \textit{deciduous} stand volume, \textit{coniferous} stand volume, stem density, and basal area. 
Model performance was evaluated using the rRMSE (\%) and the coefficient of determination ($R^{2}$), which measure prediction error and explained variance, respectively.
This work addresses three main objectives. We first investigate whether a single DL model trained on data acquired under contrasting phenological conditions (\Lon\ and \Loff) can generalize across seasons as compared to models trained with season-specific data 
, and whether this robustness holds across forest types. We then examine the specific contribution of overlapping plots, surveyed under both \Lon\ and \Loff\ conditions, as a training strategy that anchors the model to paired observations of the same forest under varying acquisition contexts. Finally, we assess the added value of auxiliary contextual variables in improving prediction accuracy across heterogeneous acquisition conditions.

\section{Materials and methods}

\subsection{Data}
Our approach perform at plot-level and is framed to handle a large diversity of context, has found in the French mainland territory. The advantages are two-fold: this ensures (i) a large amount of data and (ii) significant diversity, therefore capturing very heterogeneous behaviors necessary for our generalization task.

\subsubsection{Study area}
French forests cover approximately 17.5 million hectares, representing about 32\% of the land area of mainland France \citep{vidal_national_2016}. They are predominantly composed of broadleaved species, with around 68\% of the forest areas consisting of deciduous stands. Forest ownership is largely private, as nearly three quarters of forest land is privately owned. In terms of stand structure, more than half of French forests consist of mixed stands, combining either different broadleaved species or mixtures of broadleaved trees and conifers, while pure coniferous stands are less frequent. In metropolitan France, forests contain nearly 190 different tree species, distributed among seven main species: oak, beech, chestnut, maritime pine, Scots pine, spruce, and fir \citep{ign_memento_2025}.

This structural and compositional diversity is closely linked to the high climatic and topographic heterogeneity of France, which includes oceanic, continental, Mediterranean and mountainous climates. As a result, French forests display a wide range of forest types and structures \citep{bontemps_partition_2019, kalinicheva_super-resolved_2025}.

\subsubsection{\lidar\ data}

\begin{figure}[h]
    \centering
    \includegraphics[width=\textwidth]{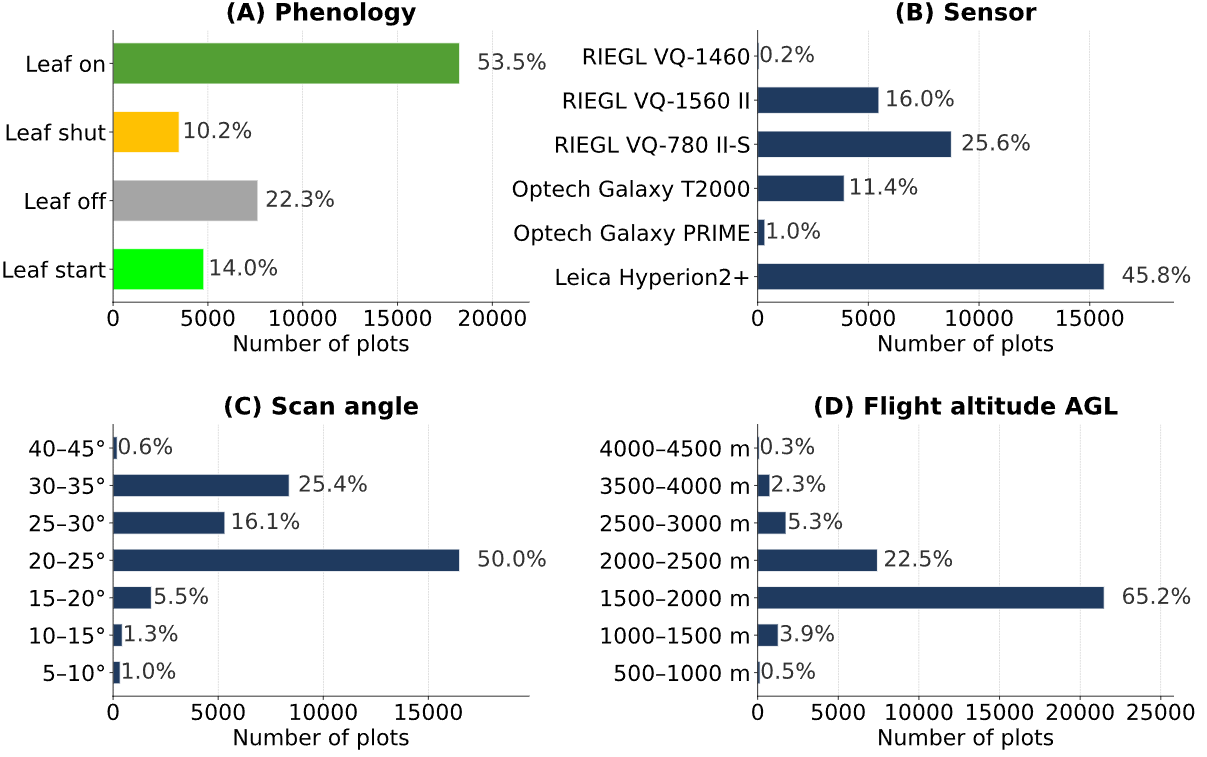}
    \caption{Variability of our plot-level \lidar\ dataset in terms of phenology, sensor, scan angle and, flight altitude. \textit{AGL} stands for \textit{Above Ground Level}.}
    \label{fig:variabilite}
\end{figure}

We benefit from the large scale 3D airborne \lidar\ data acquired in the framework of the French National \lidar\ HD program, started in 2021 for a 5$\:$years duration. The objective of the program is to acquire and release in open access \lidar\ data over the metropolitan area and ultramarine territories, with the exception of French Guyana, with a point density of at least 10$\:$pulses/m$^2$. Such a pulse density was selected to fulfill spatial sampling requirements for both Digital Terrain Model (DTM) generation, land-cover mapping and statistical tasks
\citep{gaydon_pureforest_2025}, reaching sometimes 40$\:$pts/m$^2$.
To achieve such a large multi-year campaign, the territory was divided into 50$\:$km$\times$50$\:$km patches, which were each attributed to one of the six selected private companies or to IGN. Acquisitions were conducted throughout the year, with survey parameters adapted to the relief and the \lidar\ device. Several \lidar\ systems were employed, with varying emitted energy, pulse rate and operated under multiple flight heights, plane speed and scan angle ranges (Fig.~\ref{fig:variabilite}). In addition, without inter- and intra-annual planning in the survey periods (\EG based on forest management or landscape consistency), adjacent patches may have been acquired by different companies, under different seasons, and distinct years. Edge effects between patches are reduced by integrating an overlap region of 2-5$\:$km in which at least two surveys were performed. 

Fig.~\ref{fig:variabilite} shows our dataset exhibits a significant heterogeneity in terms of season and \lidar\ surveys, leading in very distinct samplings of the tree canopy and the vegetation underneath.

Raw data processing was mainly performed using automatic routines involving TerraScan software, and classified with Myria3D DL tool \citep{gaydon_myria3d_2022}. We used the classified point cloud as the input data for our approach, as it allowed us to directly retain only the ground and vegetation classes and discard the other points, thereby simplifying the preprocessing steps.

\subsubsection{French national forest inventory data}
\label{sec:french_ifn_data}
The French NFI is based on a systematic sampling of semi-permanent plots distributed across the territory on a 1$\:$km grid. Each year, 1/10 of the grid is explored and around 14,000$\:$plots are surveyed on the field, half of them been visited for the first time, and the other half have been revisited 5$\:$years apart. Quantitative and qualitative data on standing forests and vegetation are collected in the field using concentric circular subplots of 6, 9, 15, and 25$\:$m radius. Stand description is done on the largest plots. Dendrometric data are collected on the three other radius, according to tree diameter: trees with a Diameter at Breast Height (DBH) in the range [7.5$\:$cm~–~22.5$\:$cm] are measured on the 6$\:$m radius plot, those in the range [22.5$\:$cm~$-$~47.5$\:$cm] are measured on the 9$\:$m radius plot, and larger trees on the 15$\:$m radius plot. Plot$-$level estimates of forest attributes are derived on a per $\:$ha basis using the probability of inclusion of trees, field measurements as well as allometric models for some variables like volume. In the framework of this research, we relied only on first visit plots and considered the stem density (stems ha$^{-1}$), the basal area (m$^2$ ha$^{-1}$), the stem volume (m$^3$ ha$^{-1}$) and its distribution between hardwoods and conifer, as well as dominant height (m).

\subsubsection{Our dataset}

\begin{figure}[h]
    \centering
    \includegraphics[width=0.75\textwidth]{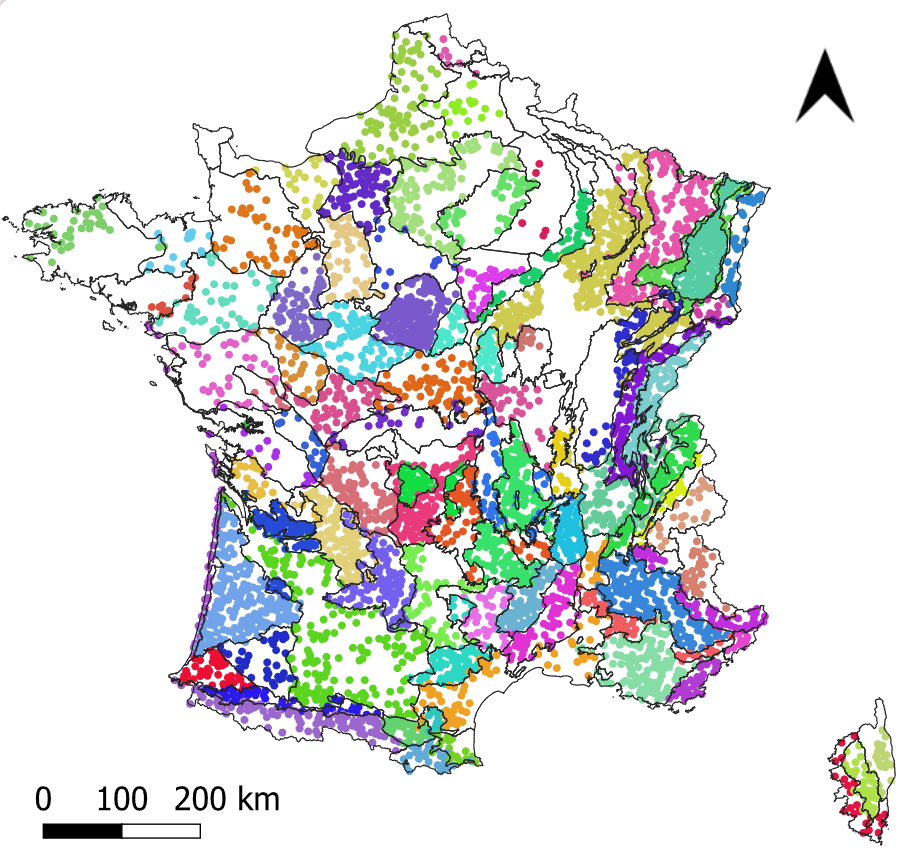}
    \caption{Spatial distribution over France of the NFI plots used in this study, colored with respect to the sylvo-ecoregion.}
    \label{fig:NFI_plots}
\end{figure}

The study focuses on the mainland territory including Corse. 80\% of the territory was covered at the time of this research. We select NFI plots with less than a five-year difference between field measurements and the \lidar\ acquisition. This results in a total of 33,877 plots (see Table \ref{tab:stats_wac} and Fig. \ref{fig:NFI_plots}), which covers approximately 7.6 million m$^2$, that sample most of the French sylvo-ecoregions (SER), French territory division based on forest ecosystems and management practices.The dataset includes 60 main species, with an average of 502.1 plots having a similar main species ($\pm$ 901.8 SD). The three most represented species are \textit{pedunculate oak} (4,166 plots), \textit{sessile oak} (3,333 plots), and \textit{European beech} (3,251 plots). We extract \lidar\ HD point clouds centered on NFI locations, and link them to the corresponding field measurements. \lidar\ point clouds were already classified and a 1$\:$m DTM made jointly available.

We used the LidR package \citep{roussel_lidr_2020} to generate Canopy Height Models (CHM) and to extract point clouds for each plot. To ensure data quality, we applied a three-step filtering procedure, retaining 32,052 plots (93.6\% of the initial dataset).
\begin{itemize}
    \item Plots with a CHM containing more than 10\% of missing values were excluded: we discard plots only partially covered by the \lidar\ acquisition.
    \item Plots with large deviations occurring between the field campaign and the \lidar\ acquisition were removed. This is likely to reflect stand disturbances, \EG harvesting or windthrow. We compared the canopy cover fraction derived from the CHM against the one recorded during field surveys as a qualitative estimate of the canopy cover by class of 10\%. Plots exhibiting a discrepancy exceeding $\pm 2\sigma$ of the observed distribution of differences were discarded. The canopy cover was defined in the \lidar\ data as the proportion of pixels with height above 2$\:$m, a threshold widely used in the literature to separate low vegetation from actual canopy returns \citep{narine_methodological_2023, rodes-blanco_canopy_2023}, and then scaled to the definition of the NFI in 10\% classes.
    \item Stand disturbances was also considered from another perspective. We compared the 95th percentile of CHM heights ($H_{95}$) to the dominant height measured in the field ($H_d$). The distribution of residuals $H_{95} - H_d$ is positively skewed: we applied an asymmetric rejection criterion: plots were excluded if $H_{95} - H_d < -2\sigma$ or $H_{95} - H_d > +3\sigma$, where $\sigma$ denotes the standard deviation of the residual distribution ($\sigma = 3.89\:$m). The stricter lower bound reflects the fact that a strong underestimation of $H_{95}$ relative to $H_d$ is most likely indicative of disturbances between the two acquisition dates, rendering the plot unsuitable for model training. Conversely, moderate overestimation of $H_{95}$ relative to $H_d$ can arise from legitimate physical sources: steep terrain slopes are known to introduce height errors in CHM normalization~\citep{vega_ptrees_2014}, and a sparse ground point density during DTM interpolation may lead to residual normalization biases, both resulting in inflated canopy height estimates. These artifacts do not necessarily reflect a mismatch between the field survey and the \lidar\ acquisition, and we wish to remain as faithful as possible to the field measurements rather than aggressively discarding potentially valid plots. Therefore, overestimation was tolerated up to a wider margin of $+3\sigma$.
\end{itemize}

\begin{table}[h]
\footnotesize
\centering
\begin{tabular}{l c c c c}
    \toprule
    \textbf{Variable} & \textbf{Min} & \textbf{Mean} & \textbf{Max} & \textbf{SD} \\
    \midrule
    Volume (m$^3$ ha$^{-1}$)  & $0.6038$  & $195.5225$ & $1655.0705$ & $159.3474$ \\
    Deciduous volume (m$^3$ ha$^{-1}$)  & $0.0000$  & $122.5009$ & $1362.7934$ & $124.9314$ \\
    Coniferous volume (m$^3$ ha$^{-1}$)   & $0.0000$  &  $73.0216$ & $1613.6613$ & $147.1187$ \\
    Density (stem ha$^{-1}$)   & $14.1092$ & $745.7198$ & $5363.3736$ & $569.1066$ \\
    Basal area (m$^2$ ha$^{-1}$)  & $0.3886$  &  $24.9230$ &  $132.0870$ &  $14.6144$ \\
    Dominant height (m)  & $2.3000$  &  $18.2193$ &   $45.5791$ &   $7.0199$ \\
    \bottomrule
\end{tabular}
\caption{Descriptive statistics of forest attributes under assessment.}
\label{tab:stats_wac}
\end{table}

\subsection{Methods}
\MaMethode\ is designed to estimate various forest attributes at the plot level using a sufficient diversity of \lidar\ point clouds and that should be agnostic to the forest type. This is a regression task that is efficiently handled with machine learning solutions and that does not require any preliminary individual tree crown detection step \citep{fan_automatic_2026}. 

\subsubsection{Model architecture}
\label{sec:architecture}
\begin{figure}[htbp] %
    \centering
    \includegraphics[width=\textwidth]{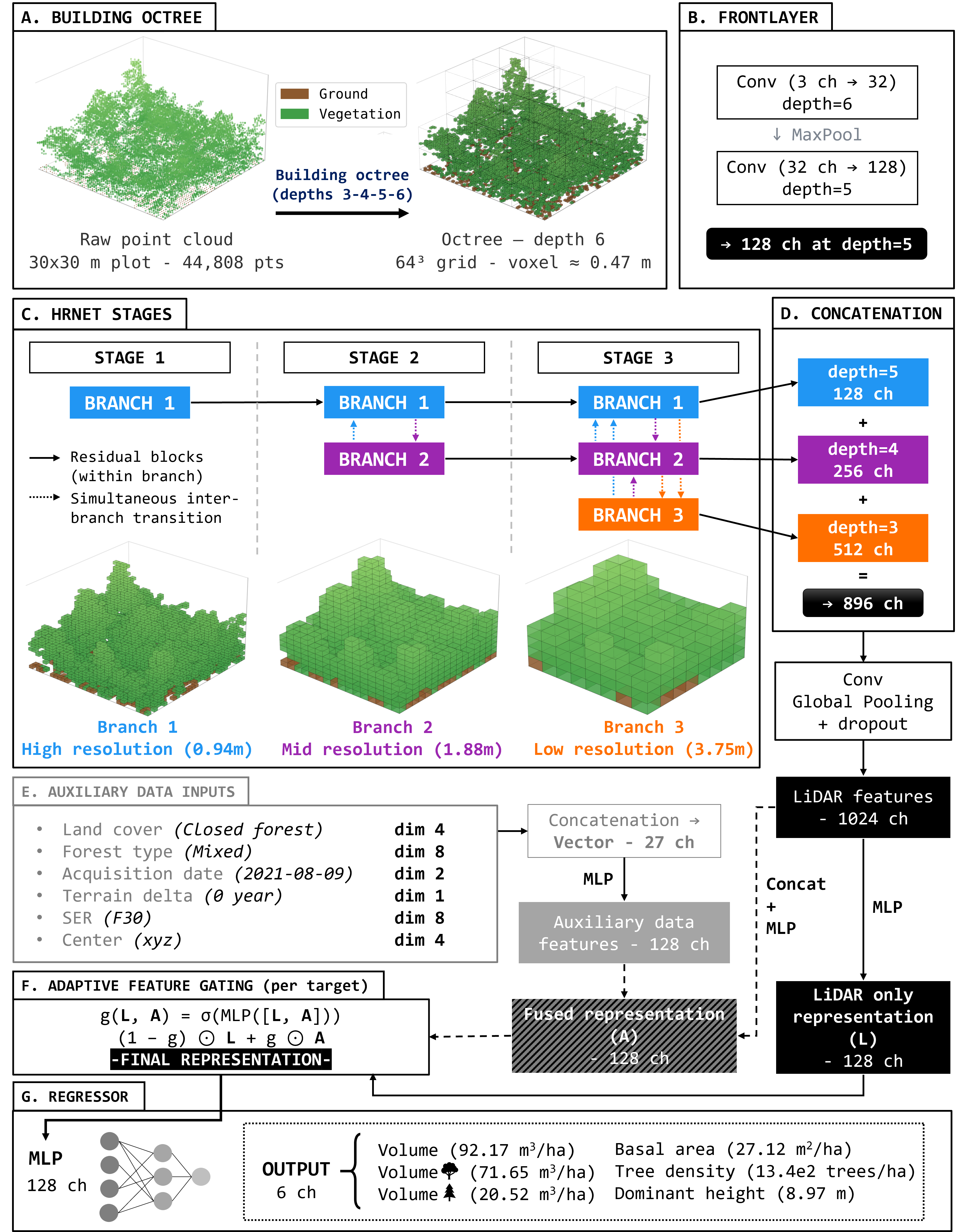} 
    \caption{Step-wise description of \MaMethode. \texttt{ch} stands for \texttt{channels}, and indicates the size of the feature space at each stage. All provided numbers correspond to the real values for the illustrated plot.}
    \label{fig:pipeline}
\end{figure}

\paragraph{Computational strategy}
Our problem can be cast as an \textit{harmonization} challenge, it should therefore exhibit a strong generalization ability over space, time, and plot composition. We selected an Octree-based Convolutional Neural Network combined with a High-Resolution Net (OCNN-HRNet) architecture developed by \citet{wang_o-cnn_2017}, and already proved effective for forest attribute regression by \citet{seely_modelling_2023}, while remaining simple.
The proposed framework is designed to operate directly on raw point cloud data, as the conversion into octree structures is automatically performed as a preprocessing step prior to model training (Fig.~\ref{fig:pipeline}).
To improve training stability and accelerate convergence, each target variable is normalized to zero mean and unit variance prior to training using a z-score transformation \citep{ayrey_synthesizing_2021}. This is motivated by the large differences in magnitude and variance across the six forest attributes (\EG volumes are expressed in m$^{3}$\,ha$^{-1}$ while basal area is in m$^{2}$\,ha$^{-1}$), which would otherwise cause the loss function to be dominated by the highest-variance attributes. Concretely, for each target $k$, the normalized value is computed as $\tilde{y}_k = (y_k - \mu_k) / \sigma_k$, where $\mu_k$ and $\sigma_k$ are the mean and standard deviation estimated over the training set. This ensures that all attributes contribute equally to the loss during training, regardless of their physical scale or absolute magnitude. Predictions are back-transformed to the original scale before evaluation.

\paragraph{Detailed design}
The model architecture begins with feature extraction from octree-based inputs, transforming raw three-dimensional point cloud data into per-node feature representations (Fig.~\ref{fig:pipeline} part A). An initial stack of octree convolution and max-pooling layers reduces the spatial resolution while increasing feature dimensionality, producing the first set of feature maps at a coarser octree depth (Fig.~\ref{fig:pipeline} part B).

The core of the network follows a HRNet architecture adapted to octree-structured data (Fig.~\ref{fig:pipeline} part C). At each stage, parallel branches process feature maps at different octree depths, corresponding to different spatial resolutions, using residual blocks with bottleneck convolutions. The first stage operates on a single high-resolution branch (128 channels). In the second stage, a second branch is introduced at half the resolution (256 channels), and in the third stage a third branch is added at an even coarser resolution (512 channels). Between stages, a transition module simultaneously exchanges information across all branch pairs: each branch receives contributions from all other branches in parallel, computed from the same input state. These contributions are resampled to the target octree depth via successive max-pooling (for higher-resolution sources) or nearest-neighbour upsampling (for lower-resolution sources), with a $1{\times}1$ convolution applied when channel dimensions differ, and summed element-wise to form the updated branch input.

After the HRNet stages, the multi-resolution feature maps are concatenated (Fig.~\ref{fig:pipeline} part D), resulting in a combined representation of 896 channels (128 + 256 + 512). A 1$\times$1 convolution and global pooling layers then aggregate spatial information into a single 1024-dimensional feature vector, with dropout applied for regularization. Finally, in the original model from \citet{seely_modelling_2023}, a multilayer perceptron regressor composed of two linear layers with ReLU activations processes this vector and outputs the final regression estimates.\\

We propose several improvements with respect to existing works on deep-based regression of forest attributes : network optimization, increased versatility and context-aware estimation. First, we perform a systematic hyperparameter search to adapt the original architecture to our prediction task. Our six targets are continuous forest structural attributes of heterogeneous scale and ecological meaning: the optimal learning rate, weight decay and batch size may differ substantially from those reported in \citep{seely_modelling_2023} that focuses on above-ground biomass estimation. Second, we design the model to rely exclusively on the 3D spatial coordinates (XYZ) of each point, deliberately excluding intensity and echo-related attributes such as return number, number of returns, and scan angle. Although these features were initially considered, hyperparameter optimization revealed no consistent performance gains from their inclusion, while intensity measurements were discarded due to the lack of radiometric sensor calibration. Restricting the input to XYZ coordinates also improves robustness across acquisition campaigns by reducing sensitivity to sensor-specific radiometric differences. Finally, we augment geometric information with contextual auxiliary data describing stand-level geographical, semantic, and temporal conditions. We hypothesize that this additional information enables the model to capture ecological and phenological significant variability that is not discernible from point cloud geometry alone. Specifically, we encode (see Fig.~\ref{fig:pipeline}):
\begin{itemize}
    \item Geographical-semantic variables (categorical, 20): canopy cover class, composition type, and sylvo-ecoregion using learnable embedding layers.
    \item Geographical knowledge (discrete, 4): we further include spatial context through a low-dimensional encoding of the plot center location.
    \item Temporal knowledge (discrete, 3): \lidar\ acquisition date is encoded with a cyclic sinusoidal representation (sine and cosine of the day of year), while the temporal gap between the \lidar\ acquisition and the field inventory date as a normalized scalar.
   
\end{itemize}
Land cover and type of forest formation are obtained from the BD For\^et\textregistered~ v2 database. BD For\^et\textregistered~ v2 is a vector database describing forest composition over France, and agregated into 32 main forest formations of pure and mixed compositions. 
Unlike field-based measurements (\EG NFI plots), which are highly accurate but spatially sparse, BD For\^et\textregistered~ offers wall-to-wall coverage suitable for large-scale modeling. The database includes a hierarchical classification that distinguishes closed and open forests according to canopy cover, as well as poplar plantations, heathlands, and herbaceous formations. Stand composition classes further differentiate pure broadleaf, pure conifer, and mixed forests.

All auxiliary data features are concatenated into a joint vector and passed through a dedicated multi-layer perceptron (MLP) to produce a learned auxiliary data representation in a shared latent space. Rather than modulating geometric features through feature-wise affine transformations like the FiLM method \citep{perez_film_2018} (unsuccessful in our case), we adopt a late fusion strategy: the auxiliary data representation is concatenated with the features extracted by the HRNet backbone and jointly processed through subsequent layers (Fig.~\ref{fig:pipeline} part E). To allow the model to adaptively control the contribution of contextual information, we introduce a gating mechanism that learns, for each prediction target, how to interpolate between geometry-driven features and fused representations (Fig.~\ref{fig:pipeline} part F).  

When the full set of auxiliary data is used, the resulting 27-dimensional representation is concatenated into a single feature vector. We adopt a late-fusion strategy, in which this auxiliary representation is combined with the geometric \lidar\ features extracted by the HRNet backbone. To allow the model to adaptively modulate the contribution of contextual information depending on the target variable, we introduce a \textbf{target-specific gating mechanism}:
\begin{equation}
\mathbf{g}_k = (1 - g_k) \odot \mathbf{L} + g_k \odot \mathbf{A},
\label{eq:gating}
\end{equation}
where $\mathbf{L} \in \mathbb{R}^{128}$ denotes the latent representation derived solely from geometric \lidar\ features, $\mathbf{A} \in \mathbb{R}^{128}$ corresponds to the fused representation combining \lidar\ and auxiliary data. $g_k = \sigma(\mathrm{MLP}([\mathbf{L}, \mathbf{A}])) \in [0,1]$ is the learned gating coefficient associated with target variable $k$. When $g_k \to 0$, the model relies exclusively on \lidar-derived geometric information, whereas when $g_k \to 1$, it fully incorporates auxiliary contextual information. This mechanism provides the model with \textbf{partial interpretability}, as the distribution of gate values on the test set enables quantification, for each forest attribute, of the relative contribution of contextual information to the final prediction. We hypothesize this design enables the network to selectively leverage auxiliary data when informative, while preserving robustness when such information noisy or less relevant.

\paragraph{Hyperparameter optimization}

DL model performance is highly sensitive to the choice of hyperparameters, such as the learning rate, batch size, and regularization coefficients. Hyperparameter tuning is therefore formulated as an optimization problem in which hyperparameters are selected to minimize a validation loss. Common approaches include grid search, random search, and Bayesian optimization \citep{raiaan_systematic_2024}. While grid search evaluates predefined combinations of hyperparameters, random search samples from predefined distributions and is generally more efficient in high-dimensional spaces \citep{Bergstra_algorithms_2011}. Bayesian optimization further improves efficiency by modeling the relationship between hyperparameters and model performance using surrogate probabilistic models.
Hyperparameter optimization was performed using Optuna \citep{akiba_optuna_2019}, an open-source framework that supports dynamic construction of the search space during execution. Optuna relies on Bayesian optimization based on a Tree-structured Parzen Estimator, which adaptively samples promising regions of the hyperparameter space. In addition, Optuna integrates pruning mechanisms that enable early stopping of unpromising trials, making it particularly suitable for computationally expensive DL models \citep{raiaan_systematic_2024, karakutuk_optuna-optimized_2025}.
The optimized hyperparameters included the learning rate, batch size, dropout probability and weight decay coefficient. These hyperparameters were selected as they have the greatest influence on optimization dynamics (learning rate, batch size) and generalization (dropout, weight decay). The search space was defined using log-uniform sampling for the learning rate and categorical sampling for the remaining parameters.

The optimization procedure was implemented in two sequential phases. First, a fixed number of trials were executed. Due to the dataset size, 50~trials required 100~hours of computation. Although a more extensive exploration of the hyperparameter space would have been desirable, the high computational cost of this procedure led us to limit the search to 50 trials. Each trial consisted of training the model for a fixed number of epochs of 10 (preliminary results showed that models converge approximately at epoch 7). Candidate configurations were ranked according to their mean validation loss. The top three configurations were then selected for a second evaluation phase. In this phase, each configuration was retrained and validated using a five-fold cross-validation scheme and a larger number of training epochs (20).
The final set of hyperparameters was chosen based on the lowest mean validation loss observed during this second phase. This two-stage strategy allows efficient exploration of the hyperparameter space while providing a robust and unbiased estimate of model generalization performance.

\subsubsection{Study Design}

The research questions outlined in Section~\ref{sec:intro} are addressed through four complementary experiments, each isolating a specific factor: the effect of phenological conditions on model generalization, the potential confound introduced by class imbalance, the role of overlapping multi-season acquisitions, and the contribution of auxiliary contextual variables together with the interpretability of their integration.

All models introduced below were built upon our \MaMethode\ architecture (see Section~\ref{sec:architecture}). They were trained in a supervised framework using paired \lidar\ point cloud samples and NFI ground truth measurement, for 20 epochs with optimized hyperparameters. Forest type was defined at the plot level from NFI records, distinguishing deciduous, coniferous, and \textit{mixed} stands. Phenological classes were defined using fixed temporal boundaries applied to acquisition dates (leaf-start: Mar.~15 ; \Lon: May~15 ; leaf-shut: Sep.~15 ; \Loff: Nov.~15): \Lon\ corresponds to full canopy closure, and \Loff\ to the leafless winter period. Transitional stages were excluded to ensure unambiguous class membership. \Lon\ and \Loff\ acquisitions represent 54\% and 22\% of the total dataset, respectively.




\paragraph{Experiment 1: All-seasons versus season-specific prediction}
We assess whether a single model trained on both phenological seasons generalizes as well as season-specific models. Three independent models were trained: a \textbf{\Lon\ model} ($n = 14{,}142$), a \textbf{\Loff\ model} ($n = 5{,}675$), and a \textbf{full model} combining both seasons ($n = 19{,}817$). Each season-specific model was evaluated under two conditions: \textbf{same-season testing} (the test set matched the training season), and \textbf{cross-season testing} (the model was applied to acquisitions from the opposite season). The full model was evaluated on both seasons. To further disentangle the effect of class distribution from that of seasonal coverage, the full model was additionally trained on a balanced dataset enforcing equal \Lon\ and \Loff\ representation through undersampling of the majority class; the \textbf{balanced model} and unbalanced original full model are compared to isolate the contribution of class imbalance (54\% \Lon\ vs.\ 22\% \Loff) from that of phenological diversity. For all configurations, samples were split into 70\% training, 10\% validation, and 20\% test, stratified by phenological class and forest type.

\paragraph{Experiment 2: Benefiting from overlapping acquisitions}

We quantify the impact of plots surveyed under both \Lon\ and \Loff\ conditions as a training strategy. Two datasets of equal size were compared: one composed exclusively of overlapping plot pairs (\IE plots with acquisitions in both seasons), and one restricted to single-acquisition plots. Overlapping plots arise from two sources: spatial overlap between adjacent acquisition blocks, and flight-line sidelap within a single block (when point clouds contained data from multiple flights, points were separated using GPS timestamps). Among the $1{,}239$ NFI plots acquired more than once across the full dataset, only $38$ were observed in both phenological seasons, yielding $76$ paired acquisitions ($38$ \Lon\ and $38$ \Loff). The single-acquisition dataset was constructed as $K = 3$ disjoint subsamples of equal size, drawn without replacement from the pool of non-overlapping plots, ensuring no plot appeared in more than one subsample. Both conditions were evaluated on a fixed holdout test set composed exclusively of single-acquisition plots. Performance is reported as mean $\pm$ standard deviation of rRMSE\% across the $K = 3$ runs.

\paragraph{Experiment 3: Assessing the contribution of auxiliary contextual information}

To assess the contribution of auxiliary data, two models were compared under otherwise identical conditions: our \textbf{full model} incorporating ecological and spatiotemporal data, fused via the late-fusion gating mechanism (\textit{Aux}), and a \textbf{base model} consisting of \MaMethode\ modified to rely solely on \lidar\ point cloud features (\textit{No Aux}). The gating coefficient $g_k \in [0,1]$ learned for each target variable $k$ quantifies the relative contribution of contextual information to the final prediction: values close to 0 indicate the model relies primarily on \lidar-derived geometric features, while values close to 1 indicate a dominant contribution from auxiliary data. Since the gates are initialized at $0.5$, values remaining close to this midpoint suggest that the auxiliary information neither contributed positively nor negatively in a substantial way. Analyzing the distribution of gate values on the test set thus provides a form of partial interpretability, enabling us to characterize, per forest attribute, how much the model draws on contextual information. This is complemented by an ablation protocol in which individual auxiliary variables are successively selected to quantify their marginal contribution. Both model variants were optimized independently using Optuna (50 trials), with the best hyperparameter configuration selected by five-fold cross-validation based on the lowest mean validation loss.

\section{Results}
\subsection{Hyperparameter tuning}

\textit{Hyperparameter optimization revealed that integrating auxiliary data fundamentally reshapes the optimization landscape, requiring a learning rate 3.5 times smaller, half the batch size, and double the dropout rate compared to the base model, while simultaneously demonstrating greater representational richness through resilience to aggressive regularization.}

This part shows the results of the hyperparameter optimization using the aforementioned setup, conducted independently for the base model and the full model. Notably, the two variants converged to markedly different configurations, which reflects the impact of auxiliary data on the model's learning dynamics. The learning rate converged to a value approximately \textbf{3.5 times lower} in the auxiliary model ($4.457 \times 10^{-4}$ vs.\ $1.570 \times 10^{-3}$), reflecting the more complex loss landscape induced by the fusion of heterogeneous information sources: the gating mechanism must learn to balance two complementary streams and is consequently more sensitive to large gradient steps. The optimal batch size likewise \textbf{halved} (16 vs.\ 32): smaller batches produce noisier gradient estimates that tend to converge towards flatter, more generalizable minima~\citep{keskar_large-batcht_2017}. This is particularly beneficial here as the auxiliary metadata introduces a combinatorial diversity of input configurations (\EG rare sylvo-ecoregion and acquisition period combinations), encouraging the model to find solutions that generalize across this expanded input space rather than overfitting to the most frequent covariate patterns. The most striking difference concerns the dropout rate, which \textbf{doubled} when auxiliary data was introduced (0.6 vs.\ 0.3): rather than viewing this solely as a defensive regularization measure, this result can be read as evidence of the representational richness brought by auxiliary data: the model remains performative despite a substantially higher fraction of masked activations, suggesting that the auxiliary features provide sufficiently redundant and complementary signals for the network to tolerate aggressive information dropout, while simultaneously preventing the model from short-circuiting point cloud processing by treating high-level contextual features, such as forest type, geographic region, or acquisition date, as direct proxies for the target variable. Conversely, weight decay was \textbf{reduced by a factor of approximately three} ($3.6 \times 10^{-5}$ vs.\ $1 \times 10^{-4}$): since the stronger dropout already provides substantial regularization, a lower $\ell_2$ penalty avoids over-constraining the learnable embeddings associated with categorical metadata, which require sufficient parametric freedom to encode meaningful distinctions between categories. Taken together, these results paint a coherent picture of how auxiliary data reshapes the optimization problem: it substantially enriches the representational space, as evidenced by the model's resilience to aggressive dropout, while the added complexity of multimodal fusion calls for more conservative gradient updates, and the combination of smaller batches and reduced weight decay ensures that this richer representation generalizes effectively across the full diversity of input configurations.

\subsection{All-seasons versus season-specific prediction}

\subsubsection{Results for full model}

\begin{table}[H]
\centering
\small
\begin{tabular}{lcc cc cc}
\toprule
& \multicolumn{2}{c}{\textbf{\ULon~test}}
& \multicolumn{2}{c}{\textbf{\ULoff~test}}
& \multicolumn{2}{c}{\textbf{All test}} \\
\cmidrule(lr){2-3}
\cmidrule(lr){4-5}
\cmidrule(lr){6-7}
\textbf{Metric} & rRMSE & R$^2$ & rRMSE & R$^2$ & rRMSE & R$^2$ \\
\midrule
Volume       & 40.83 & 0.76 & 36.21 & 0.67 & 39.28 & 0.736 \\
Conif. Vol.  & 53.95 & 0.75 & 43.09 & 0.69 & 49.72 & 0.742 \\
Decid. Vol.  & 77.23 & 0.82 & 96.25 & 0.83 & 81.99 & 0.821 \\
Density      & 53.72 & 0.47 & 57.44 & 0.42 & 54.82 & 0.460 \\
Basal Area   & 36.81 & 0.56 & 34.03 & 0.48 & 36.03 & 0.544 \\
Height       & 12.81 & 0.89 & 11.44 & 0.81 & 12.33 & 0.884 \\
\bottomrule
\end{tabular}
\caption{\textbf{Full model tested separately on \Lon, \Loff, and complete datasets}; results for the three forest types (Coniferous, Deciduous and Mixed; see details in \ref{app:full_model_results}).}
\label{tab:full-model}
\end{table}

\textit{The full model achieves strong performance on height and total volume, while stem density remains the most challenging attribute (R$^2$ < 0.47), with species-specific volume predictions exhibiting high variability across forest types.}

Table \ref{tab:full-model} presents the global performance of the full model across all forest plots, evaluated on both \Lon\ and \Loff\ test sets. Details for forest types are provided in \ref{app:full_model_results}.

Height is the best-predicted attribute by a large margin, with error below 13\% and R$^2$$\,>\,$0.89 on \Lon, and remains stable across forest types. Total volume also yields strong results (error $\approx$ 40\%, R$^2$ = 0.76 on \Lon), with consistent performance across forest types. Basal area achieves moderate accuracy (error $\approx$ 36\%, R$^2$ = 0.56 on \Lon), while stem density is the most poorly predicted attribute, with R$^2$ below 0.47 on both test sets.

Predictions for the proportion of deciduous and conifers exhibit the highest errors and the widest per-forest-type variability. Coniferous and deciduous volumes reach errors of 53.95\% and 77.23\% on \Lon\ respectively, with extreme rRMSE values on ecologically mismatched forest types (\EG deciduous volume on coniferous plots), where target values are near zero. R$^2$ values nonetheless remain relatively high for deciduous volume (0.82--0.83), reflecting good rank-order predictions despite high absolute errors.

With the exception of density, and volume of deciduous, other attributes \Loff\ errors are generally lower than or comparable to \Lon, while R$^2$ values tend to be higher on \Lon. This trend indicates that \Lon\ capture better their overall variance, but is impacted by larger prediction errors.

When evaluated on the combined \textit{All test} dataset, performance remains highly consistent with the separate \Lon\ and \Loff\  evaluations, with only marginal changes in both rRMSE and $R^2$ across all targets. Overall trends observed in the separate test sets are preserved, with height and total volume remaining the most accurately predicted, while stem density and species-specific volume fractions continue to be the most challenging targets.

\subsubsection{Balanced}

\begin{figure}[H] %
    \centering
    \includegraphics[width=\textwidth]{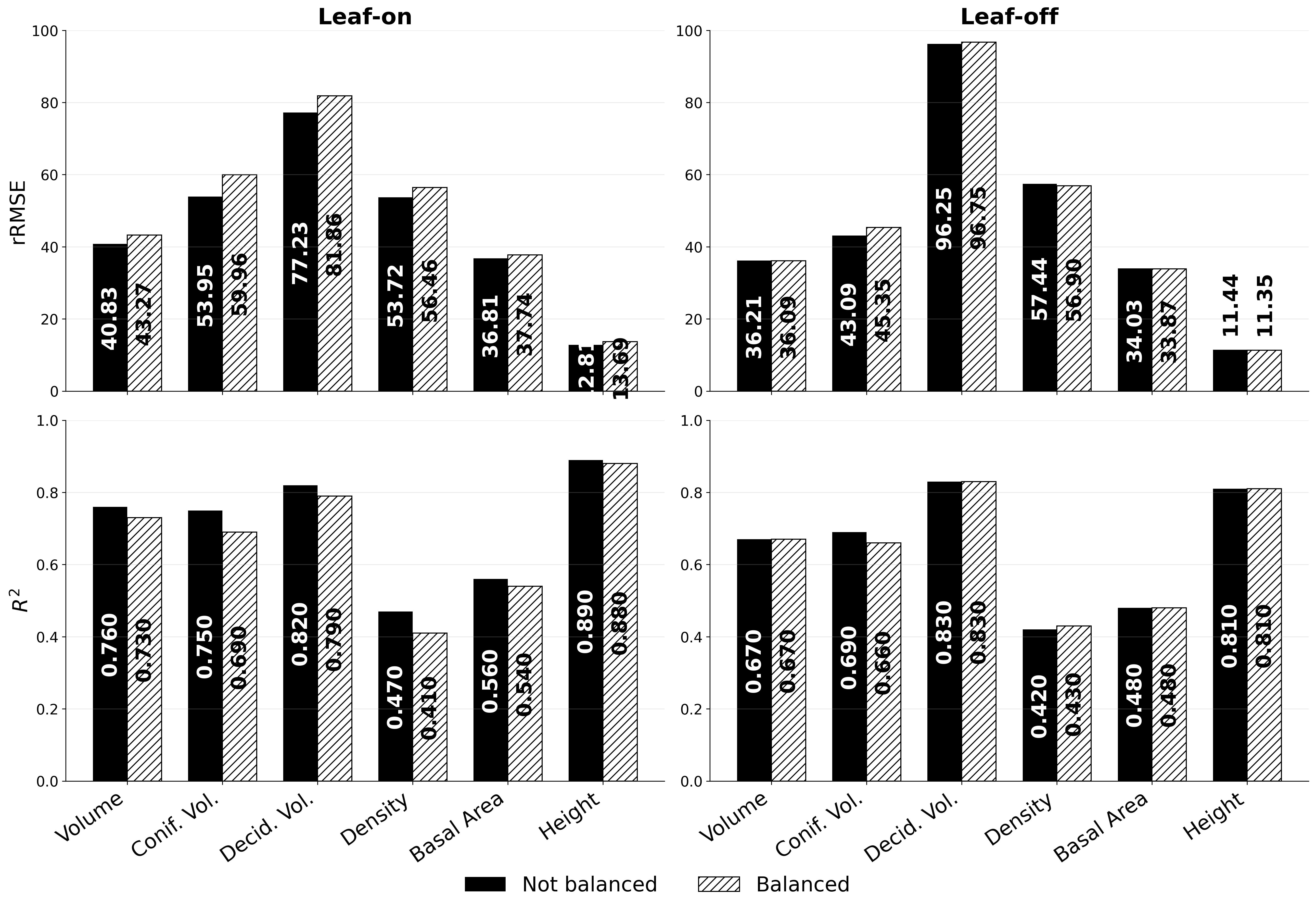} 
    \caption{
    Comparison of predictive performance between the full model (solid black bars) and the balanced model (hatched bars) across all target forest attributes. Results are reported for rRMSE (\%) and $R^2$ on \Lon\ and \Loff\ test sets.
    }
    \label{fig:balanced}
\end{figure}

\textit{Balancing the training set leads to moderate performance degradation on \Lon\ while largely preserving \Loff\ performance, suggesting that the original training imbalance does not critically bias the full model.}

Figure \ref{fig:balanced} compares the full and balanced models across all target attributes. Balancing the training set leads to a moderate degradation of \Lon\ performance across most attributes, while \Loff\ performance is largely preserved or marginally improved. The effect is most pronounced for coniferous volume and stem density, where \Lon\ error increases by 6 and 3 points respectively, and for height, where \Lon\ R$^2$ drops from 0.890 to 0.880. In contrast, height and basal area show the smallest sensitivity to the balancing strategy, with differences below 1 point of error on both test sets. Overall, the performance gap between balanced and unbalanced models remains limited, suggesting that the initial imbalance does not critically bias the full model. Detailed per-forest-type results are reported in \ref{app:balanced}.

\subsubsection{Season-Specific Models}

\begin{table}[H]
\centering
\small
\begin{tabular}{lcccccccc}
\toprule
 & \multicolumn{4}{c}{\textbf{\ULon~model}} & \multicolumn{4}{c}{\textbf{\ULoff~model}} \\
\cmidrule(lr){2-5}\cmidrule(lr){6-9}
\textbf{Metric} & \multicolumn{2}{c}{Same-season} & \multicolumn{2}{c}{Cross-season} & \multicolumn{2}{c}{Same-season} & \multicolumn{2}{c}{Cross-season} \\
\cmidrule(lr){2-3}\cmidrule(lr){4-5}\cmidrule(lr){6-7}\cmidrule(lr){8-9}
 & $\Delta$rRMSE & $\Delta$R$^2$ & $\Delta$rRMSE & $\Delta$R$^2$ & $\Delta$rRMSE & $\Delta$R$^2$ & $\Delta$rRMSE & $\Delta$R$^2$ \\
\midrule
Volume      & +0.97 & -0.01 & +3.22 & -0.07 & +3.46 & -0.07 & +24.32 & -0.38 \\
Conif. Vol. & +3.54 & -0.03 & +7.24 & -0.11 & +10.94 & -0.18 & +45.13 & -0.38 \\
Decid. Vol. & +6.87 & -0.04 & +33.15 & -0.14 & +22.52 & -0.09 & +119.06 & -1.01 \\
Density     & +0.43 & -0.01 & +5.93 & -0.13 & +1.23 & -0.03 & +14.07 & -0.32 \\
Basal Area  & +0.98 & -0.02 & +3.74 & -0.12 & +1.76 & -0.06 & +19.47 & -0.40 \\
Height      & +1.17 & -0.02 & +0.78 & -0.03 & +4.64 & -0.18 & +5.79 & -0.03 \\
\bottomrule
\end{tabular}
\caption{Difference in performance between season-specific and full models.
$\Delta$RMSE = rRMSE(seasonal) $-$ rRMSE(full); $\Delta$R$^2$ = R$^2$(seasonal) $-$ R$^2$(full),
where the full model reference is always evaluated on the same test set as the season-specific model.}
\label{tab:season-specific}
\end{table}

\textit{Season-specific models consistently underperform the full model even within their calibration domain, with cross-season generalization exhibiting strong asymmetry: the \Lon\ model transfers reasonably well to \Loff\  data, while the \Loff\ model degrades severely when applied to \Lon\ acquisitions.}

Table~\ref{tab:season-specific} compares season-specific and full model performance under same-season and cross-season evaluation. {The full model has better performance than the seasonal one when $\Delta$rRMSE is positive and $\Delta$R$^2$ negative.

\paragraph{Same-season performance}
Models trained on either \Lon\ or \Loff\ data show slightly lower performances than the full one for both rRMSE and R$^2$ (Tab.~\ref{tab:season-specific}). This trend applies to both leaf conditions, indicating that the full model performs better than the specific ones, even in their calibration domain. For \Lon, differences remain modest: volume error increases by less than 1 point (41.80\% vs. 40.83\%), and stem density and basal area stay within 1\% of the full model. The degradation is more visible for coniferous volume (57.49\% vs. 53.95\%) and deciduous volume (84.10\% vs. 77.23\%).

On \Loff, the gap widens. While basal area and stem density remain comparable (35.79\% vs. 34.03\% and 58.67\% vs. 57.44\%), deciduous volume shows a pronounced degradation (118.77\% vs. 96.25\%) and height drops substantially in both error and explained variance (16.08\%, R$^2$\,=\,0.626 vs. 11.44\%, R$^2$\,=\,0.811), suggesting that training on a single season is particularly limiting for structurally sensitive attributes.

\paragraph{Cross-season performance}
Cross-season generalization is strongly asymmetric between the two seasonal models. On one hand, the \Lon\ model applied to \Loff\ acquisitions maintains acceptable performance for most attributes: The rRMSE of volume increases by 3.22\% and R$^2$ decreases by 0.07, staying close to same-season performance, and height remains robust with less than 1\% error difference. The main degradation in error is observed for deciduous volume (+33.15\%) and coniferous volume (+7.24\%). On the other hand, the \Loff\ model applied to \Lon\ acquisitions degrades severely across all attributes. Volume rRMSE increases by 24.32\%, coniferous volume by 45.13\%, and deciduous volume collapses by 119.06\%. Basal area R$^2$ drops by 0.4 and stem density R$^2$ by 0.32, while height, though degraded in error (+5.79\%), partially retains its R$^2$ (-0.03). This asymmetry suggests that \Loff\ acquisitions carry less transferable structural information than \Lon\ data when applied outside their training distribution.

\paragraph{Comparison with the full model}
The full model consistently outperforms season-specific models in both settings. In same-season testing, its advantage is clearest on \Loff, where it reduces volume error by more than 3 points and nearly halves the height error gap (11.44\% vs. 16.08\%, R$^2$\,=\,0.811 vs. 0.626). In cross-season testing, the full model avoids the severe degradation of the \Loff-specific model: it maintains volume error at 40.83\% on \Lon where the \Loff-specific model reaches 65.15\%, and preserves R$^2$ above 0.75 for volume and height across both test sets. per-forest-type results are reported in~\ref{app:seasonal_model_results}.

\subsection{Can we benefit from overlapping acquisitions ?}

\begin{figure}[H] %
    \centering
    \includegraphics[width=\textwidth]{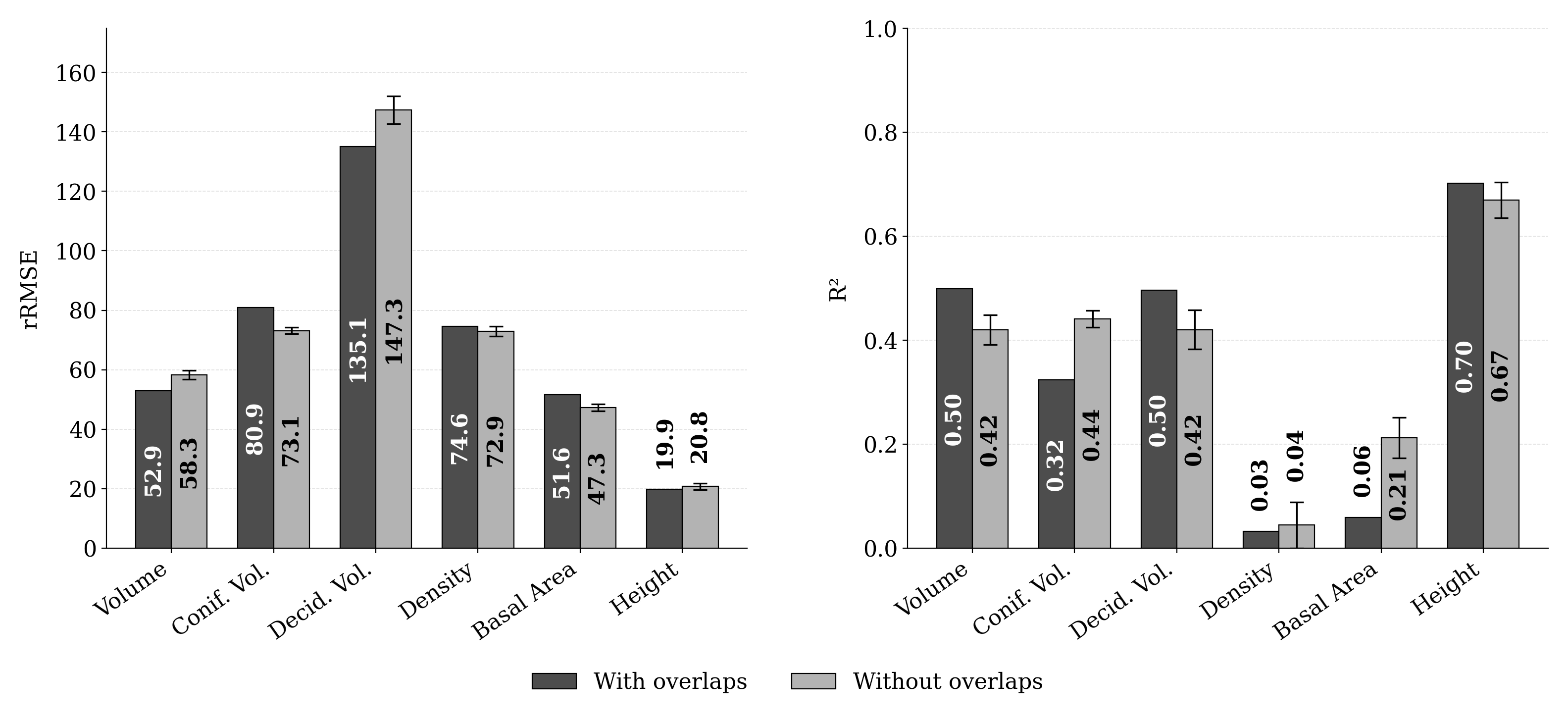} 
    \caption{
    Effect of spatial overlap between \lidar\ acquisitions on model performance.
    Metrics (rRMSE and R\textsuperscript{2}) are compared for overlapping vs. non-overlapping areas,
    averaged across all species. Values for non-overlapping areas represent mean $\pm$ standard deviation over $K$ runs.
    }
    \label{fig:overlap}
\end{figure}

\textit{Spatial overlap between \lidar\ acquisitions produces contrasting effects depending on the target attribute: volumetric estimation improves substantially for total and deciduous volume, while coniferous volume, basal area, and density degrade, and height remains largely unaffected.}

Overlapping acquisitions yield contrasting effects depending on the target attribute (Figure~\ref{fig:overlap} and~\ref{app:overlap}). Performance improves most clearly for total and deciduous volume, with rRMSE reductions of roughly 5 and 12 points respectively, suggesting that the complementary seasonal information captured by paired acquisitions benefits volumetric estimation of broadleaf stands. Coniferous volume shows the opposite trend, with rRMSE increasing from $73.15\%$ to $80.87\%$ and R\textsuperscript{2} dropping from $0.441$ to $0.324$. Basal area and density also degrade under the overlap condition, while height remains largely unaffected in either direction. Compared to the full model, results in this experiment are strongly degraded for all target variables (\EG $\approx$+10 points in volume error) probably due to the reduced train dataset size.

\subsection{Do auxiliary contextual information bring additional knowledge ?}

\textit{Auxiliary data yield limited improvements across most attributes, with species-specific volumes showing the highest gate activations and largest rRMSE reductions, while height demonstrates no improvement linked to auxiliary data and density systematically degrades.}

\begin{figure}[H] %
    \centering
    \includegraphics[width=\textwidth]{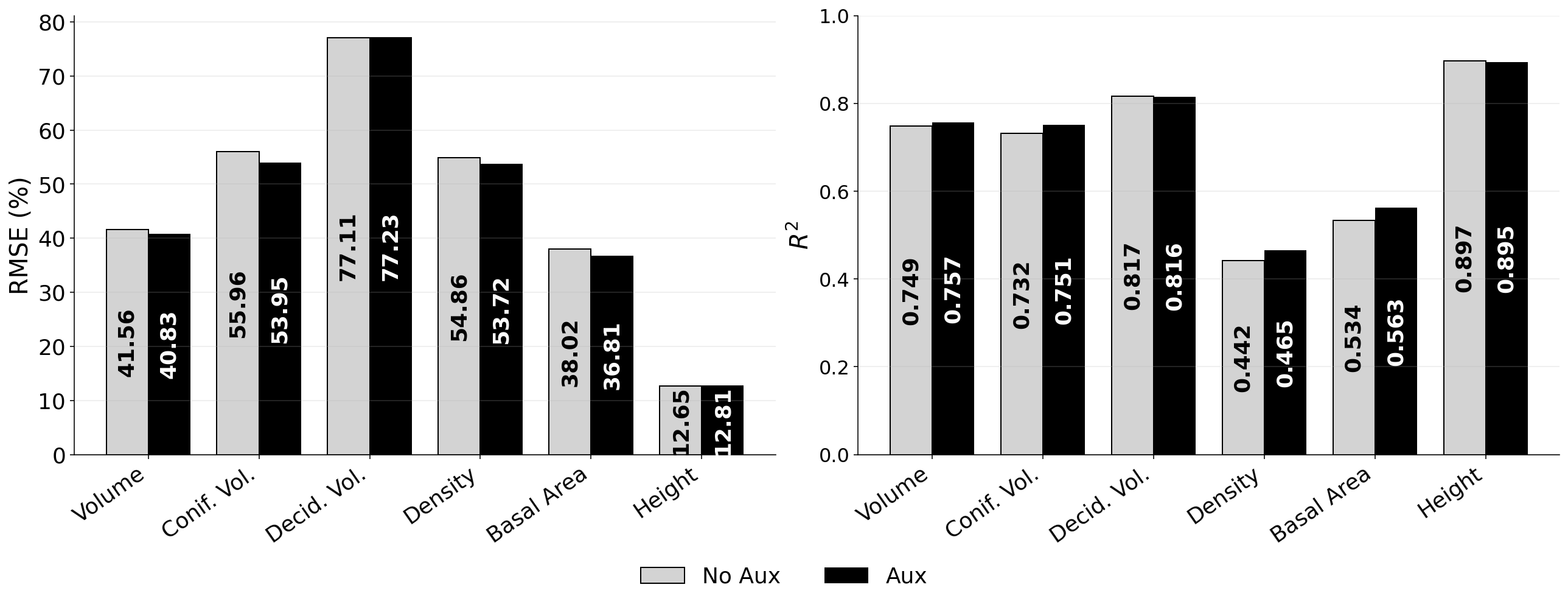} 
    \caption{Barplots showing changes in rRMSE and R$^2$ between the \lidar-only baseline (No Aux) and \MaMethode\ (Aux).}
    \label{fig:barplots_aux_no_aux}
\end{figure}%

\paragraph{Impact on predictive accuracy (Figure~\ref{fig:barplots_aux_no_aux})}
Figure~\ref{fig:barplots_aux_no_aux} compares global rRMSE and R$^2$ obtained with \lidar-only inputs (grey bars) to the \MaMethode\ architecture including auxiliary data (black bars), both trained on the full training set. Detailed numerical results are provided in~\ref{app:full_model_results}. Across attributes, changes remain moderate in absolute magnitude, but clear patterns emerge. For \textbf{total volume}, rRMSE decreases by approximately 0.7\%, while R$^2$ remains stable. Coniferous volume shows slightly larger effects, with rRMSE reductions of 2\% and R$^2$ gains reaching +0.02 while performance on deciduous volume are stable. Basal area (rRMSE -1.21, R$^2$ +0.03) and Density (rRMSE -1.14, R$^2$ +0.02) exhibit small improvements. In contrast, height show minimal sensitivity to auxiliary data: rRMSE variations remain under 0.2\%, with R$^2$ changes below 0.01. Overall, auxiliary data lead to limited improvements, but more noticeable for species-specific volume components.

\begin{figure}[H] %
    \centering
    \includegraphics[width=\textwidth]{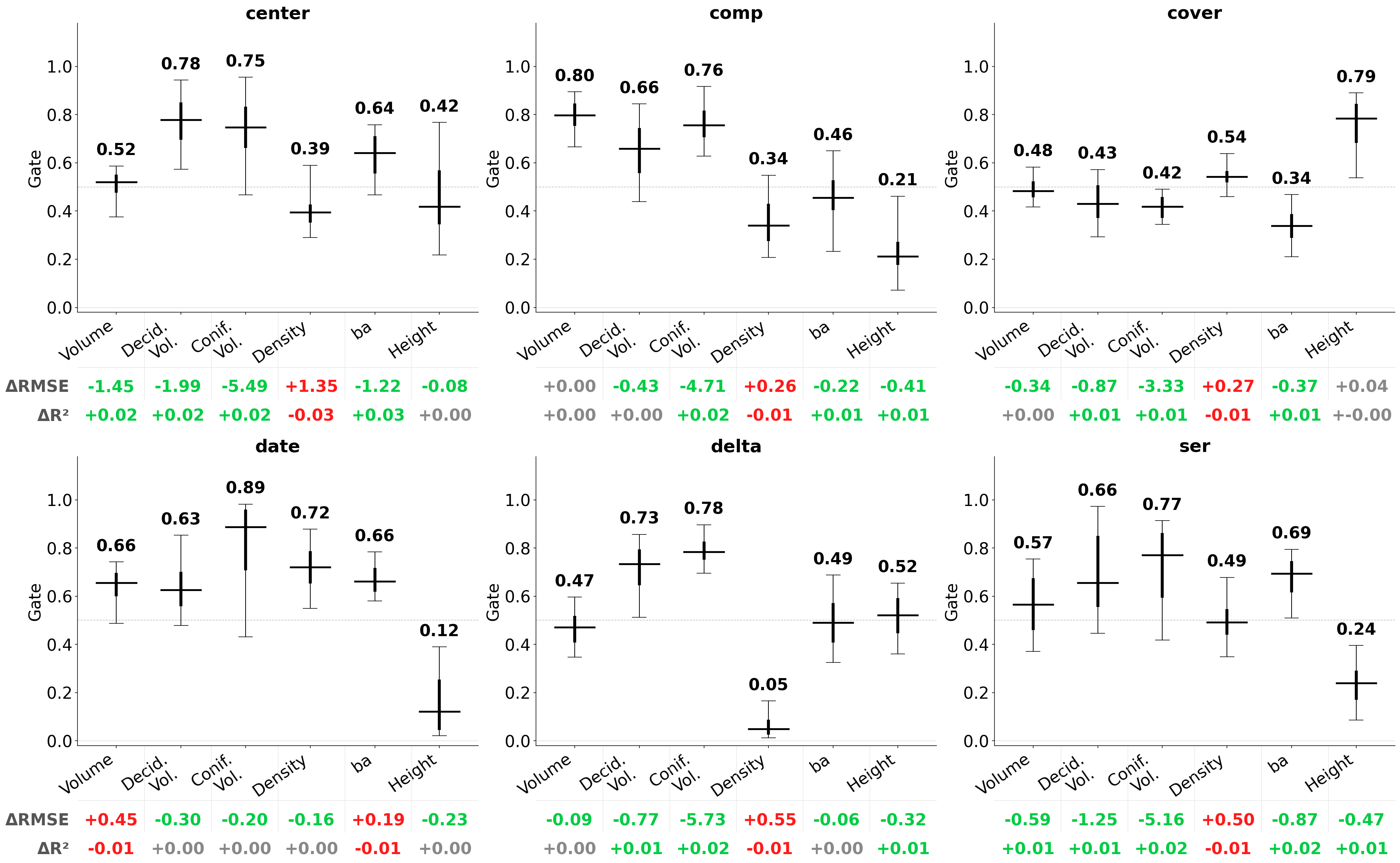} 
    \caption{
    Distribution of learned gating coefficients for each auxiliary data configuration ($\mathrm{Config})$ across target forest attributes. 
    For each configuration, whiskers represent the 5th--95th percentile range, thick vertical segments indicate the interquartile range, and horizontal black lines denote median gate values. 
    The dashed horizontal line marks $g=0.5$. 
    Values reported below each panel show relative performance differences with respect to the \lidar-only baseline ($\Delta = \mathrm{Config} - \mathrm{No\_Aux}$), expressed as changes in rRMSE (\%) and $R^2$. 
    Green and red values respectively denote positive and negative variations according to the corresponding metric convention.
    }
    \label{fig:meta_ablation_gates}
\end{figure}

Figure~\ref{fig:meta_ablation_gates} reports the median gate values and associated performance differences for each auxiliary variable and target attribute. Full results are reported in~\ref{app:Ecological}.

Species-specific volumes benefit most from auxiliary information. Coniferous volume consistently exhibits the highest gate activations across all configurations ($0.66$--$0.89$), regardless of which variable is provided, with rRMSE reductions reaching $-5.73$ for the time \texttt{delta} between \lidar\ and field acquisition alone and $-6.20$ for the full auxiliary set. Deciduous volume follows a similar pattern, with gate values between $0.63$ and $0.78$ and consistent improvements up to $-1.99$ points for \texttt{center}. Total volume and basal area show intermediate activations ($\approx$$0.47$--$0.66$) and more modest but generally positive contributions. Height displays the most heterogeneous gating behavior, ranging from 0.12 (\texttt{date}) to 0.79 (\texttt{cover}), yet without consistent performance benefit, high gate values do not systematically translate into lower error, suggesting the model gates toward auxiliary data without effectively exploiting it for this attribute. Density is the only attribute for which auxiliary information systematically degrades performance across configurations, with rRMSE increases up to $+1.35$ for \texttt{center}, despite gate values remaining near 0.5.

Among individual variables, \texttt{center} and \texttt{delta} yield the strongest marginal contributions for volume-related attributes, while \texttt{date} and \texttt{cover} have negligible or inconsistent impact across most targets.
\section{Discussion}

\subsection{Full model performance}

The full model achieves competitive results for height and total volume, which are the best-predicted attributes across all configurations (rRMSE\,$<$\,13\% and R$^2$\,$>$\,0.88 for height; rRMSE\,$\approx$\,39\% and R$^2$\,=\,0.74 for total volume on the combined test set). For height, these results are consistent with recent studies using ALS data at national scale in Sweden: \citet{nilsson_nationwide_2017} report stand-level rRMSE of 9.8--11.2\% using linear regression on 11,500 NFI plots, and \citet{bjornberg_improving_2026} report approximately 10--12\% using XGBoost and CNN on NFI plots. The differences on total volume is more pronounced: at the local scale, \citet{hawrylo_use_2020} report rRMSE of 24.16--31.69\% over 360 stands within a single Polish forest district using ALS combined with Landsat composites and various machine learning methods, while at the national scale, \citet{nilsson_nationwide_2017} report rRMSE of 17.2--22.0\% and \citet{bjornberg_improving_2026} approximately 30\%. 
Both papers were conducted on Swedish boreal forests where Norway spruce and Scots pine together account for approximately 80\% of standing volume. Conversely, our dataset comprises 33,877 NFI plots spanning 60 main species with $\approx$\,72\% of plots dominated by deciduous stands, making national-scale prediction substantially more challenging.

This spatial contrast is further supported by \citet{miettinen_pan-european_2025}, who show strong regional variability in Sentinel-2 kNN-based forest estimates over 151,000 NFI plots across 14 European countries divided in multiple ecological zones. In the area including Swedish boreal reference region, model performance is consistently higher than in most other areas, with rRMSE of 58.2\% for volume, and moderate predictive skill (R$^2$\,=\,0.54). In contrast, areas covering the French territory exhibit systematically higher errors, with volume rRMSE ranging from 61.7\% to 74.6\% confirming the impact of forest structure and composition on performance.

For basal area, our global results (rRMSE$\approx$\,36.0\%, R$^2$\,=\,0.54 on the combined test set) align reasonably with recent literature. \citet{keskes_improving_2025} in Romania report, at 10$\:$m resolution, an R$^2$ of 0.28--0.35 and rRMSE of 21.7--32.3\%. Over Sweden, \citet{nilsson_nationwide_2017} achieve rRMSE ranging between 20.4--26.7\%. At an even larger scale, \citet{matasci_large-area_2018} mapped structural attributes across the ~552\,million hectares of Canadian boreal forest using 30$\:$m annual Landsat composites (1984--2012 time series) combined with \lidar\ plots from a national transect survey and a kNN-Random Forest imputation approach, reporting R$^2$\,=\,0.544 and rRMSE of 50.7\% for basal area. Our results, obtained with national scale high-density \lidar, thus compare favourably with large-area mapping studies relying on coarser remotely sensed data, while remaining above the accuracy levels reported for local or compositionally simpler settings. 

Stem density is the most poorly predicted attribute across all configurations (RMSE~$\approx$~55\%, R$^2$~$<$~0.47), a result consistent with the broader literature. For example, in a subtropical planted forest in southern China, \citet{liu_deep_2021} found that stem density was among the most challenging forest structural parameters to estimate from airborne \lidar\ data, despite using a hybrid deep-learning model (Deep-RBN) that outperformed conventional approaches. Stem density predictions exhibited substantially lower accuracy than estimates of tree height and volume, with an rRMSE =20.34\%--31.03\% and R$^2$\,=\,0.52–0.72 under the optimal model configuration. In the two studies focusing on high-density ALS in New Zealand plantation forests of \textit{radiata pine}, stem density consistently yields the lowest performance among all forest attributes considered. Using standard ALS metric-based regression models, \citet{pearse_comparison_2018} report a R$^2$ of 0.48 and rRMSE of 34\% for stand density, compared to 0.88 for mean top height, noting explicitly that ALS models produced poor estimates of stand density. Similarly, using Random Forest regression, \citet{leonardo_comparison_2020}, in the southeast of New Zealand, achieve an R$^2$ of 0.47 and rRMSE of 32.3\% for stem density, versus 0.86 for height. Unlike height or volume, stem density is strongly influenced by stand history, thinning regime, and understorey structure-information that is poorly encoded in plot-level point cloud geometry and that auxiliary data cannot fully compensate for, as further evidenced by \citet{irulappa-pillai-vijayakumar_increasing_2019}, who attribute the comparatively poor density predictions of their k-NN model to the structural complexity of broadleaved-dominated forests in central France.

To our knowledge, no existing study has directly predicted absolute coniferous and deciduous volumes as separate target variables. The closest comparable work is that of \citet{miettinen_pan-european_2025}, who produced Pan-European maps of deciduous-coniferous proportion (DCP). While their DCP metric provides the proportion of conifers (allowing broadleaf proportion to be derived as 100--DCP), the absolute volumes of each species group remain unquantified. In the French territory covered by their study, the rRMSE for conifer proportion ranged from 61.8\% to 71.6\% on the processing area corresponding to France. This finding underscores the inherent difficulty of species-specific attribution. Our results for coniferous volume (RMSE $\approx$ 50\%, R$^2$\,= \,0.74) and deciduous volume (RMSE $\approx$ 82\%, R$^2$\,= \,0.82) showcase the challenging task of predicting absolute species-specific volumes. Despite the high error values, the R$^2$ metrics remain reasonably strong, suggesting that the models capture spatial variability and stand composition patterns well despite substantial prediction errors at the plot level.

\subsection{Effect of balanced training}

Balancing the training set to equalize the representation of \Lon\ and \Loff\ acquisitions leads to a moderate and selective degradation of performance. On \Lon\ test data, most attributes show slightly higher errors under the balanced configuration, most notably for coniferous volume (+6 points rRMSE) and stem density (+3 points), while \Loff\ performance is largely preserved or marginally improved (Table in ~\ref{tab:full_model_results_balanced}). Height and basal area show the smallest sensitivity to balancing, with differences below 1 point on both test sets (Figure~\ref{fig:balanced}). This pattern is consistent with the composition of the original dataset: \Lon\ acquisitions represent the majority of plots, so balancing effectively down-samples the dominant season and reduces the model's exposure to the most frequently occurring acquisition context. The limited overall impact suggests that the original imbalance does not critically bias the full model, and that the unbalanced configuration is preferable for operational use given its better exploitation of available training data. This is consistent with findings in class-imbalanced learning more broadly, where down-sampling the majority class tends to improve minority-class performance but at the cost of overall accuracy~\citep{japkowicz_class_2002}

\subsection{Phenological robustness and season-specific generalization}

The results presented in Tables~\ref{tab:season-specific} and \ref{app:seasonal_model_results} address two complementary questions: whether season-specific models outperform the full model on their own acquisition season, and whether they generalize adequately to the opposite season.

\subsubsection{Same-season performance and the advantage of leaf-off}

In same-season evaluation, the \Loff\ model achieves lower rRMSE than the \Lon\ model for most volumetric attributes (\EG total volume: 39.67\% vs. 41.80\%), consistent with the literature showing that \Loff\ acquisitions generally yield more accurate volume estimates due to greater canopy penetration and improved ground sampling 
\citep{villikka_suitability_2012, white_evaluating_2015}. This advantage is however attribute-dependent: height estimation degrades substantially under \Loff\ same-season conditions (16.08\% vs. 13.98\% for the \Lon\  model), likely because the absence of foliage reduces the contrast between canopy strata and lowers the upper return density that drives height predictions. This is consistent with findings by \citet{davison_effect_2020}, who show that \Loff\ conditions shift the return distribution toward lower canopy levels, particularly in deciduous stands.

The \Loff\ model also shows marked degradation for deciduous volume in same-season evaluation, with rRMSE dropping from 118.77\% to values exceeding 150\% without auxiliary data. This reflects a well-known difficulty: under \Loff\ conditions, mixed stands containing both deciduous and evergreen species present ambiguous return distributions, as retained conifer foliage introduces structural noise that complicates the separation of species-specific contributions \citep{villikka_suitability_2012}. \citet{white_evaluating_2015} report similar patterns, noting that \Loff\ models show larger errors for deciduous volume attributes, particularly for merchantable volume, where rRMSE can differ by up to 7\% between leaf conditions.

\subsubsection{Cross-season generalization and asymmetric transferability}

Cross-season generalization is strongly asymmetric. The \Lon\ model applied to \Loff\ acquisitions maintains acceptable performance for most attributes: total volume reaches 39.43\% rRMSE (R$^2$\,=\,0.60), close to its same-season performance, and height remains robust at 12.22\% (R$^2$\,=\,0.78). By contrast, the \Loff\ model applied to \Lon\ acquisitions degrades severely: total volume increases to 65.15\% (R$^2$\,=\,0.38), coniferous volume to 85.96\%, and deciduous volume collapses to 196.29\% (R$^2$\,=\,$-$0.19). This asymmetry is consistent with \citet{white_evaluating_2015}, who report that applying \Loff\ models to \Lon\ data on eight forest attributes (top height, mean height, Lorey's mean height, basal area, quadratic mean diameter, merchantable volume, total volume, and total aboveground biomass) results in average absolute rRMSE increases of 16.8\% for coniferous stands and 25.8\% for deciduous stands, and with \citet{villikka_suitability_2012}, who observe that a \Loff\ model applied to \Lon\ deciduous plots yields rRMSE exceeding 70\% and bias approaching 50\%. The physical explanation is straightforward: a model trained on \Loff\ point clouds, where returns are concentrated in the lower canopy and on the ground, lacks the representational capacity to interpret the denser and higher return distributions, representative of \Lon\ acquisitions \citep{orka_effects_2010}.

\subsubsection{Practical implications for acquisition planning}

These results raise a practical question for operational forest inventory: when is it preferable to acquire \Lon\ versus \Loff\ data? The answer depends on the intended application. \ULoff\ acquisitions are generally preferred for terrain modeling and archaeological prospection, where reduced foliage improves ground return detection and the visibility of micro-topographic features beneath forest canopies \citep{simpson_assessment_2017, moudry_comparison_2019}. Conversely, \Lon\ acquisitions provide a more realistic characterization of canopy fuels and vegetation continuity, making them more suitable for wildfire-related mapping applications \citep{moran_mapping_2020}. 
For multi-attribute forest inventory, our results suggest that \Lon\ acquisitions are preferable when cross-season generalization is required, given their superior transferability to \Loff\ conditions. However, acquisition season is not always a free parameter: mountainous areas are often only accessible under \Lon\ conditions due to snow cover precluding \Loff\ surveys. Regulatory constraints can further restrict the acquisition window in some regions. These operational realities motivate the development of models capable of handling heterogeneous acquisition conditions, precisely the challenge that the full \MaMethode\ model addresses.

\subsubsection{The full model as a robust operational solution}

The full model, trained on both \Lon\ and \Loff\ data, consistently outperforms season-specific models regardless of the test set phenology. In same-season evaluation, its advantage is clearest on \Loff, where it reduces total volume rRMSE from 39.67\% to 36.21\% and nearly halves the height error gap (11.44\% vs. 16.08\%, R$^2$\,=\,0.81 vs. 0.63). In cross-season setting, it avoids the severe degradation of the \Loff-specific model, maintaining volume rRMSE at 40.83\% on \Lon\ ,where the \Loff\ model reaches 65.15\%. This is consistent with \citet{white_evaluating_2015}, who show that pooled models trained on both seasons achieve intermediate performance between \Lon\ and \Loff\ models in same-season evaluation, while substantially outperforming cross-season model-mixing. Our results extend this finding to a DL setting at the national scale, suggesting that training on heterogeneous phenological conditions is not only feasible but produces more robust predictions than any season-specific strategy, a practically important result given the uneven and fragmented nature of \Lon/\Loff\ coverage across the French national \lidar~programme.

\subsection{Contribution of overlapping acquisitions}

Overlapping plots \IE~surveyed under both \Lon\ and \Loff\ conditions, provide a unique training signal by anchoring the model to paired observations of the same forest under contrasting contexts. The results in Table~\ref{tab:overlap-results} and Figure~\ref{fig:overlap} show that exploiting these overlaps yields contrasting effects, depending on the target attribute.
Performance improves most clearly for total and deciduous volume, with rRMSE reductions of approximately 5 and 12 points respectively. This suggests that paired phenological observations help the model better disentangle broadleaf structural signals across seasons. Height also shows modest but consistent improvements under the overlap condition. These gains are coherent with the premise that co-registered multi-seasonal acquisitions provide complementary canopy penetration profiles that are informative for broadleaf volume estimation \citep{orka_effects_2010, davison_effect_2020}.

Coniferous volume shows the opposite trend, with rRMSE increasing from 73.15\% to 80.87\% and R$^2$ dropping from 0.441 to 0.324 under the overlap condition. Basal area and stem density also degrade slightly. This result is counterintuitive but may reflect the limited size and geographic scope of the overlap subset: overlapping areas represent a geographically constrained fraction of the territory, and the model trained on these plots may lack sufficient diversity in coniferous stand types to generalize. Furthermore, coniferous stands are structurally more stable across phenological conditions than deciduous stands \citep{wasser_influence_2013}, meaning that paired \Lon/\Loff\ observations provide comparatively less complementary structural information for conifers. As a result, restricting training to overlap plots likely reduces effective sample diversity, without providing the compensating benefit of paired phenological contrast that drives improvements for broadleaf volume estimation. This suggests that exploiting dense optical time series (\EG Landsat or Sentinel-2) can help mitigate these limitations by capturing phenological variability that single-date or acquisition-tied observations cannot resolve. Time series approaches improve the discrimination of forest types by separating phenology-driven spectral signals from structural variability, which is particularly relevant in mixed or conifer-dominated stands where seasonal contrast is weaker and less informative \citep{guindon_new_2024}.

\subsection{Added value of auxiliary variables}

The relevance of contextual variables was evaluated both through direct Aux (\MaMethode)/No Aux comparisons (\ref{tab:full_model_results} and~\ref{tab:LOFF_model_full_results}) and through ablation gating analysis (Figure~\ref{fig:meta_ablation_gates}). Together, these results reveal that auxiliary metadata provides limited improvements in overall and same‑season conditions, but substantial gains in cross‑season generalization, with strong contrasts across forest attributes and forest‑type strata.

The improvements become much larger in cross‑season settings. Both season‑specific models exhibit strong error inflation when tested on the opposite season, and in these cases auxiliary metadata plays a major stabilizing role. For example, the \Loff\ model tested on \Lon\ exhibits rRMSE reductions from 97.7\% to 65.1\% for total volume, from 121.9\% to 85.9\% for coniferous volume, and from 367\% to 196\% for deciduous volume. Similar effects are observed for the \Lon\ model cross‑tested on \Loff, with deciduous volume rRMSE improving from 168.8\% to 129.4\%. These results confirm that temporal and ecological metadata compensate for seasonal structural shifts—especially in \Loff\ acquisitions where canopy geometry is less informative.

\textbf{Coniferous volume} and \textbf{deciduous volume} benefit the most from auxiliary information, consistent with their low geometric discriminability in the point cloud. Ablation gates identify \texttt{delta}, \texttt{center} and \texttt{SER} as the most influential variables, and the Aux/No Aux comparisons confirm this through systematic rRMSE and R$^2$ gains for pure coniferous and pure deciduous stands. This is coherent with the strong spatial segregation of species types (\EG coniferous stands at higher elevation), which the contextual variables make explicit. In contrast, mixed stands often degrade when auxiliary data is added, suggesting that ecological metadata pushes the model towards over‑interpreting coarse species patterns that do not reflect the finer structural mosaics of mixed forests.

\textbf{Total volume and basal area} show moderate gate activations and correspondingly modest but stable improvements, particularly through the \texttt{center} and \texttt{ser} variables. These two variables capture regional trends in productivity and forest structure that \lidar\ alone cannot fully encode. Their effects remain positive across most models and test conditions.

\textbf{Height} is the only attribute essentially unaffected by auxiliary metadata. Gate activations are heterogeneous but do not translate into performance gains/ The Aux/No Aux comparisons across all tables show changes below 0.2\% rRMSE. This confirms that height is almost entirely determined by \lidar\ canopy geometry and largely independent of ecological descriptors.

\textbf{Stem density} exhibits a more complex pattern. In same‑season tests, auxiliary variables mildly degrade performance, consistent with shortcut learning dynamics where categorical metadata encodes coarse density priors that interfere with the structurally weak \lidar\ signal. However, in cross‑season tests, the effect reverses: density rRMSE improves substantially (\EG 82.1\% → 67.8\% for the \Loff\ model tested on \Lon). Thus, while metadata can be detrimental in homogeneous acquisition contexts, it provides valuable stabilization when acquisition conditions shift, mirroring the behaviour observed for species‑specific volumes.

Overall, \texttt{center}, \texttt{ser}, and \texttt{delta} stand out as the variables providing the most reliable improvements—particularly for volume‑related attributes and in cross‑season scenarios—while \texttt{date} and \texttt{cover} contribute little or inconsistently. The convergence of direct performance comparisons and gating analysis highlights that auxiliary variables are not universally beneficial but become crucial whenever \lidar\ geometry alone is insufficient to capture ecological variability or to bridge strong shifts in acquisition conditions.

\section{Conclusion}

This study introduced \MaMethode, a voxel-based DL framework for predicting six forest structural attributes from heterogeneous airborne \lidar\ data at national scale, trained and evaluated on 32,052 NFI plots across mainland France.

The main findings are as follows. A single model trained jointly on \Lon\ and \Loff\ acquisitions consistently outperforms season-specific models, including within their respective calibration domains, and avoids the strong cross-season degradation observed when \Loff\ models are applied to \Lon\ data. The initial class imbalance between \Lon\ and \Loff\ samples does not bias the full model, and using all available training data proves preferable to enforcing class balance. Plots acquired under both phenological conditions improve estimates of broadleaf and total volume, while their benefit for coniferous stands is limited. Auxiliary contextual variables provide modest but consistent gains, particularly for species-specific volumes and cross-season generalization, whereas dominant height is driven almost entirely by point cloud geometry.

Overall, these results show that a generic DL approach trained on sufficiently large and diverse national-scale data can handle acquisition heterogeneity such as that of the French \lidar\ HD without explicit stratification by season, sensor, or forest type. The model also exhibits strong generalization, an increasingly critical requirement for large-scale applications where \lidar\ acquisitions are spatially heterogeneous, calling for methods capable of robust performance over extensive areas. Species-specific volume prediction remains the main limitation, highlighting the need to integrate dense optical time series to better capture phenological variation and enable operational wall-to-wall mapping.


\section{CRediT authorship contribution statement}

\noindent\textbf{Emilie Vautier:} Writing – original draft, Writing – review and editing, Data Curation, Conceptualization, Investigation, Methodology, Formal Analysis, Software, Validation, Visualization\\
\textbf{Clément Mallet:} Writing – review and editing, Conceptualization, Methodology, Formal Analysis, Supervision\\
\textbf{Cédric Vega:} Writing – review and editing, Conceptualization, Methodology, Formal Analysis, Supervision\\

\section{Declaration of Competing Interest}
\noindent The authors declare that there are no conflicts of interest associated with this work.

\appendix

\section{Detailed results of \MaMethode\ trained on the full training set}
\label{app:full_model_results}
\begin{table}[H]
\label{tab:full_model_results}
\centering
\scriptsize
\sisetup{round-mode=places, round-precision=2}
\setlength{\tabcolsep}{4pt}
\begin{tabular}{l l S[round-mode=places,round-precision=2,table-format=4.2] S[round-mode=places,round-precision=2,table-format=1.3,table-sign-mantissa] S[round-mode=places,round-precision=2,table-format=4.2] S[round-mode=places,round-precision=2,table-format=1.3,table-sign-mantissa] S[round-mode=places,round-precision=2,table-format=4.2] S[round-mode=places,round-precision=2,table-format=1.3,table-sign-mantissa] S[round-mode=places,round-precision=2,table-format=4.2] S[round-mode=places,round-precision=2,table-format=1.3,table-sign-mantissa]}
\toprule
& &
\multicolumn{4}{c}{\textbf{\Lon\ testing}} &
\multicolumn{4}{c}{\textbf{\Loff\ testing}} \\
\cmidrule(lr){3-6}\cmidrule(lr){7-10}
\textbf{Metric / Species} &
& \multicolumn{2}{c}{No Aux} & \multicolumn{2}{c}{\MaMethode}& \multicolumn{2}{c}{No Aux} & \multicolumn{2}{c}{\MaMethode} \\
\cmidrule(lr){3-4}\cmidrule(lr){5-6}\cmidrule(lr){7-8}\cmidrule(lr){9-10}
& 
& {rRMSE} & {R\textsuperscript{2}} & {rRMSE} & {R\textsuperscript{2}}& {rRMSE} & {R\textsuperscript{2}} & {rRMSE} & {R\textsuperscript{2}} \\
\midrule
\multicolumn{10}{l}{\textbf{Full model}} \\
\midrule
\multirow{4}{*}{\textbf{Volume}} & All & 41.5600 & 0.74877 & 40.8285\text{$\uparrow$} & \bfseries 0.75753\text{$\uparrow$} & 36.2257 & 0.66544 & \bfseries 36.2072\text{$\uparrow$} & 0.66579\text{$\uparrow$} \\
 & Coniferous & 42.6647 & 0.63925 & 40.2303\text{$\uparrow$} & \bfseries 0.67924\text{$\uparrow$} & 37.2105 & 0.62245 & \bfseries 36.8397\text{$\uparrow$} & 0.62994\text{$\uparrow$} \\
 & Deciduous & 40.2357 & 0.80753 & 37.9540\text{$\uparrow$} & \bfseries 0.82874\text{$\uparrow$} & 30.9505 & 0.68107 & \bfseries 30.2264\text{$\uparrow$} & 0.69582\text{$\uparrow$} \\
 & Mixed & 38.8493 & \bfseries 0.72909 & 44.8795\text{$\downarrow$} & 0.63847\text{$\downarrow$} & \bfseries 36.7080 & 0.69458 & 39.9351\text{$\downarrow$} & 0.63852\text{$\downarrow$} \\
\midrule
\multirow{4}{*}{\textbf{Conif. Vol.}} & All & 55.9567 & 0.73233 & 53.9510\text{$\uparrow$} & \bfseries 0.75118\text{$\uparrow$} & \bfseries 42.8657 & 0.69248 & 43.0937\text{$\downarrow$} & 0.68920\text{$\downarrow$} \\
 & Coniferous & 43.7677 & 0.63380 & 41.7672\text{$\uparrow$} & \bfseries 0.66651\text{$\uparrow$} & 38.7186 & 0.60822 & \bfseries 38.2649\text{$\uparrow$} & 0.61735\text{$\uparrow$} \\
 & Deciduous & 324.4291 & -0.32518 & 262.1770\text{$\uparrow$} & \bfseries 0.13459\text{$\uparrow$} & 245.6485 & -1.15953 & \bfseries 218.8721\text{$\uparrow$} & -0.71440\text{$\uparrow$} \\
 & Mixed & 78.6949 & 0.43770 & 83.3641\text{$\downarrow$} & 0.36899\text{$\downarrow$} & \bfseries 58.5740 & \bfseries 0.46831 & 75.8378\text{$\downarrow$} & 0.10871\text{$\downarrow$} \\
\midrule
\multirow{4}{*}{\textbf{Decid. Vol.}} & All & \bfseries 77.1109 & 0.81657 & 77.2341\text{$\downarrow$} & 0.81599\text{$\downarrow$} & 96.1060 & \bfseries 0.82802 & 96.2454\text{$\downarrow$} & 0.82753\text{$\downarrow$} \\
 & Coniferous & 500.2732 & -1.19530 & \bfseries 469.3694\text{$\uparrow$} & -0.93245\text{$\uparrow$} & 632.9520 & -1.22832 & 562.2492\text{$\uparrow$} & \bfseries -0.75830\text{$\uparrow$} \\
 & Deciduous & 43.0763 & 0.78620 & 40.5516\text{$\uparrow$} & \bfseries 0.81053\text{$\uparrow$} & 35.5642 & 0.60736 & \bfseries 35.4216\text{$\uparrow$} & 0.61050\text{$\uparrow$} \\
 & Mixed & 62.0253 & 0.51032 & 71.3068\text{$\downarrow$} & 0.35280\text{$\downarrow$} & \bfseries 59.4789 & \bfseries 0.55234 & 64.7828\text{$\downarrow$} & 0.46895\text{$\downarrow$} \\
\midrule
\multirow{4}{*}{\textbf{Density}} & All & 54.8580 & 0.44239 & \bfseries 53.7249\text{$\uparrow$} & \bfseries 0.46519\text{$\uparrow$} & 55.3382 & 0.45971 & 57.4385\text{$\downarrow$} & 0.41792\text{$\downarrow$} \\
 & Coniferous & 56.0820 & 0.42759 & \bfseries 54.7682\text{$\uparrow$} & 0.45409\text{$\uparrow$} & 55.2540 & \bfseries 0.46622 & 56.9276\text{$\downarrow$} & 0.43339\text{$\downarrow$} \\
 & Deciduous & 49.9650 & 0.47049 & \bfseries 49.7662\text{$\uparrow$} & \bfseries 0.47469\text{$\uparrow$} & 57.9706 & 0.43996 & 59.5359\text{$\downarrow$} & 0.40931\text{$\downarrow$} \\
 & Mixed & 50.8873 & 0.33394 & \bfseries 49.6896\text{$\uparrow$} & 0.36492\text{$\uparrow$} & 52.2326 & \bfseries 0.42727 & 58.2358\text{$\downarrow$} & 0.28805\text{$\downarrow$} \\
\midrule
\multirow{4}{*}{\textbf{Basal Area}} & All & 38.0221 & 0.53373 & 36.8077\text{$\uparrow$} & \bfseries 0.56304\text{$\uparrow$} & \bfseries 33.4660 & 0.49399 & 34.0264\text{$\downarrow$} & 0.47690\text{$\downarrow$} \\
 & Coniferous & 39.7444 & 0.40762 & 37.8716\text{$\uparrow$} & 0.46213\text{$\uparrow$} & 34.7076 & 0.45237 & \bfseries 34.3896\text{$\uparrow$} & \bfseries 0.46236\text{$\uparrow$} \\
 & Deciduous & 35.4189 & 0.65779 & 33.4854\text{$\uparrow$} & \bfseries 0.69413\text{$\uparrow$} & \bfseries 27.7137 & 0.43423 & 29.6792\text{$\downarrow$} & 0.35113\text{$\downarrow$} \\
 & Mixed & 36.7988 & 0.55470 & 38.6875\text{$\downarrow$} & 0.50782\text{$\downarrow$} & \bfseries 33.0420 & \bfseries 0.56923 & 37.1188\text{$\downarrow$} & 0.45638\text{$\downarrow$} \\
\midrule
\multirow{4}{*}{\textbf{Height}} & All & 12.6528 & \bfseries 0.89742 & 12.8116\text{$\downarrow$} & 0.89483\text{$\downarrow$} & \bfseries 11.2339 & 0.81752 & 11.4401\text{$\downarrow$} & 0.81075\text{$\downarrow$} \\
 & Coniferous & 13.0044 & 0.88464 & 12.8292\text{$\uparrow$} & \bfseries 0.88772\text{$\uparrow$} & \bfseries 11.3526 & 0.80250 & 11.6696\text{$\downarrow$} & 0.79132\text{$\downarrow$} \\
 & Deciduous & 12.3932 & \bfseries 0.91810 & 12.7917\text{$\downarrow$} & 0.91274\text{$\downarrow$} & \bfseries 10.2802 & 0.87499 & 10.2975\text{$\downarrow$} & 0.87457\text{$\downarrow$} \\
 & Mixed & 11.7811 & \bfseries 0.89905 & 12.7758\text{$\downarrow$} & 0.88128\text{$\downarrow$} & 11.4893 & 0.83396 & \bfseries 10.9760\text{$\uparrow$} & 0.84846\text{$\uparrow$} \\
\bottomrule
\end{tabular}
\caption{The full model, trained on both \Loff\ and \Lon\ samples, is evaluated on \Lon\ and \Loff\ test sets. Performance is reported in terms of rRMSE and R\textsuperscript{2} for both \MaMethode\ and No Aux configurations. Arrows on \MaMethode\ values indicate whether \MaMethode\ outperforms No Aux ($\uparrow$ \MaMethode\ better, $\downarrow$ No Aux better).}
\end{table}

\section{Detailed results of \MaMethode\ comparing balanced and non-balanced full training sets}
\label{app:balanced}

\begin{table}[H]
\label{tab:LON_model_results_balanced}
\centering
\scriptsize
\sisetup{round-mode=places, round-precision=2}
\setlength{\tabcolsep}{4pt}
\begin{tabular}{l l S[round-mode=places,round-precision=2,table-format=4.2] S[round-mode=places,round-precision=2,table-format=1.3,table-sign-mantissa] S[round-mode=places,round-precision=2,table-format=4.2] S[round-mode=places,round-precision=2,table-format=1.3,table-sign-mantissa] S[round-mode=places,round-precision=2,table-format=4.2] S[round-mode=places,round-precision=2,table-format=1.3,table-sign-mantissa] S[round-mode=places,round-precision=2,table-format=4.2] S[round-mode=places,round-precision=2,table-format=1.3,table-sign-mantissa]}
\toprule
& &
\multicolumn{4}{c}{\textbf{Same-season testing}} &
\multicolumn{4}{c}{\textbf{Cross-season testing}} \\
\cmidrule(lr){3-6}\cmidrule(lr){7-10}
\textbf{Metric / Species} &
& \multicolumn{2}{c}{No Aux} & \multicolumn{2}{c}{\MaMethode}& \multicolumn{2}{c}{No Aux} & \multicolumn{2}{c}{\MaMethode} \\
\cmidrule(lr){3-4}\cmidrule(lr){5-6}\cmidrule(lr){7-8}\cmidrule(lr){9-10}
& 
& {rRMSE} & {R\textsuperscript{2}} & {rRMSE} & {R\textsuperscript{2}}& {rRMSE} & {R\textsuperscript{2}} & {rRMSE} & {R\textsuperscript{2}} \\
\midrule
\multicolumn{10}{l}{\textbf{\Lon\ model}} \\
\midrule
\multirow{4}{*}{\textbf{Volume}} & All & 43.2248 & \bfseries 0.72824 & 45.1909\text{$\downarrow$} & 0.70295\text{$\downarrow$} & 41.2675 & 0.56584 & \bfseries 40.8694\text{$\uparrow$} & 0.57418\text{$\uparrow$} \\
 & Coniferous & 43.9871 & \bfseries 0.61654 & 45.6708\text{$\downarrow$} & 0.58662\text{$\downarrow$} & \bfseries 42.0745 & 0.51729 & 42.7162\text{$\downarrow$} & 0.50246\text{$\downarrow$} \\
 & Deciduous & 40.7756 & \bfseries 0.80233 & 42.1692\text{$\downarrow$} & 0.78859\text{$\downarrow$} & 35.8774 & 0.57145 & \bfseries 31.6556\text{$\uparrow$} & 0.66637\text{$\uparrow$} \\
 & Mixed & 43.4686 & \bfseries 0.66084 & 47.0106\text{$\downarrow$} & 0.60332\text{$\downarrow$} & 42.6031 & 0.58861 & \bfseries 42.4444\text{$\uparrow$} & 0.59167\text{$\uparrow$} \\
\midrule
\multirow{4}{*}{\textbf{Conif. Vol.}} & All & 59.9212 & \bfseries 0.69306 & 64.0260\text{$\downarrow$} & 0.64957\text{$\downarrow$} & 55.2607 & 0.48893 & \bfseries 52.5100\text{$\uparrow$} & 0.53854\text{$\uparrow$} \\
 & Coniferous & \bfseries 45.5006 & \bfseries 0.60422 & 48.6042\text{$\downarrow$} & 0.54839\text{$\downarrow$} & 50.1229 & 0.34345 & 47.4624\text{$\uparrow$} & 0.41130\text{$\uparrow$} \\
 & Deciduous & 428.5160 & -1.31189 & 412.1722\text{$\uparrow$} & -1.13890\text{$\uparrow$} & 258.4921 & -1.39126 & \bfseries 239.0392\text{$\uparrow$} & \bfseries -1.04489\text{$\uparrow$} \\
 & Mixed & 90.0083 & \bfseries 0.26440 & 100.1783\text{$\downarrow$} & 0.08878\text{$\downarrow$} & \bfseries 77.5509 & 0.06799 & 78.1627\text{$\downarrow$} & 0.05323\text{$\downarrow$} \\
\midrule
\multirow{4}{*}{\textbf{Decid. Vol.}} & All & \bfseries 80.3934 & \bfseries 0.80062 & 87.9130\text{$\downarrow$} & 0.76158\text{$\downarrow$} & 147.9330 & 0.59253 & 132.6112\text{$\uparrow$} & 0.67256\text{$\uparrow$} \\
 & Coniferous & \bfseries 451.9093 & \bfseries -0.79136 & 542.1549\text{$\downarrow$} & -1.57826\text{$\downarrow$} & 1598.7323 & -13.21633 & 1248.5789\text{$\uparrow$} & -7.67097\text{$\uparrow$} \\
 & Deciduous & 45.0604 & \bfseries 0.76605 & 47.3827\text{$\downarrow$} & 0.74132\text{$\downarrow$} & \bfseries 39.8910 & 0.50601 & 42.0416\text{$\downarrow$} & 0.45130\text{$\downarrow$} \\
 & Mixed & \bfseries 68.6364 & \bfseries 0.40036 & 77.5168\text{$\downarrow$} & 0.23516\text{$\downarrow$} & 70.8505 & 0.36481 & 69.0972\text{$\uparrow$} & 0.39586\text{$\uparrow$} \\
\midrule
\multirow{4}{*}{\textbf{Density}} & All & 58.9171 & 0.35682 & \bfseries 57.9150\text{$\uparrow$} & \bfseries 0.37851\text{$\uparrow$} & 65.1703 & 0.25067 & 67.1538\text{$\downarrow$} & 0.20436\text{$\downarrow$} \\
 & Coniferous & 60.6956 & 0.32953 & \bfseries 58.9206\text{$\uparrow$} & \bfseries 0.36817\text{$\uparrow$} & 65.3041 & 0.25438 & 66.8951\text{$\downarrow$} & 0.21760\text{$\downarrow$} \\
 & Deciduous & \bfseries 53.6503 & \bfseries 0.38950 & 54.6804\text{$\downarrow$} & 0.36583\text{$\downarrow$} & 65.6622 & 0.28149 & 70.0035\text{$\downarrow$} & 0.18334\text{$\downarrow$} \\
 & Mixed & \bfseries 52.0108 & \bfseries 0.30420 & 52.9462\text{$\downarrow$} & 0.27895\text{$\downarrow$} & 63.3222 & 0.15826 & 65.0647\text{$\downarrow$} & 0.11129\text{$\downarrow$} \\
\midrule
\multirow{4}{*}{\textbf{Basal Area}} & All & 39.4769 & \bfseries 0.49737 & 40.3833\text{$\downarrow$} & 0.47402\text{$\downarrow$} & \bfseries 38.3134 & 0.33678 & 41.0559\text{$\downarrow$} & 0.23844\text{$\downarrow$} \\
 & Coniferous & 41.1420 & \bfseries 0.36523 & 41.7734\text{$\downarrow$} & 0.34559\text{$\downarrow$} & \bfseries 39.7440 & 0.28191 & 42.6039\text{$\downarrow$} & 0.17484\text{$\downarrow$} \\
 & Deciduous & 36.2271 & \bfseries 0.64199 & 36.6923\text{$\downarrow$} & 0.63274\text{$\downarrow$} & \bfseries 31.4142 & 0.27305 & 33.5248\text{$\downarrow$} & 0.17208\text{$\downarrow$} \\
 & Mixed & 39.4780 & \bfseries 0.48750 & 41.8784\text{$\downarrow$} & 0.42328\text{$\downarrow$} & \bfseries 38.2444 & 0.42291 & 41.0881\text{$\downarrow$} & 0.33389\text{$\downarrow$} \\
\midrule
\multirow{4}{*}{\textbf{Height}} & All & 15.8742 & 0.83854 & 14.9739\text{$\uparrow$} & \bfseries 0.85634\text{$\uparrow$} & 12.6272 & 0.76944 & \bfseries 12.3992\text{$\uparrow$} & 0.77769\text{$\uparrow$} \\
 & Coniferous & 16.3066 & 0.81861 & 14.9971\text{$\uparrow$} & \bfseries 0.84657\text{$\uparrow$} & 12.7752 & 0.74990 & \bfseries 12.7337\text{$\uparrow$} & 0.75152\text{$\uparrow$} \\
 & Deciduous & 15.4256 & \bfseries 0.87311 & 15.6558\text{$\downarrow$} & 0.86930\text{$\downarrow$} & 12.4527 & 0.81658 & \bfseries 11.2437\text{$\uparrow$} & 0.85046\text{$\uparrow$} \\
 & Mixed & 15.0203 & 0.83590 & 13.7053\text{$\uparrow$} & \bfseries 0.86338\text{$\uparrow$} & 11.5539 & 0.83208 & \bfseries 10.9749\text{$\uparrow$} & 0.84849\text{$\uparrow$} \\
\bottomrule
\end{tabular}
\caption{The \Lon\ model, trained exclusively on \Lon\ samples, is evaluated on a same-season test set (\Lon) and cross-tested on the \Loff\ season. Performance is reported in rRMSE and R\textsuperscript{2} for both \MaMethode\ (with auxiliary data) and No Aux configurations. Arrows on \MaMethode\ values indicate whether \MaMethode\ outperforms No Aux ($\uparrow$ \MaMethode\ better, $\downarrow$ No Aux better).}
\end{table}

\begin{table}[H]
\label{tab:full_model_results_balanced}
\centering
\scriptsize
\sisetup{round-mode=places, round-precision=2}
\setlength{\tabcolsep}{4pt}
\begin{tabular}{l l S[round-mode=places,round-precision=2,table-format=4.2] S[round-mode=places,round-precision=2,table-format=1.3,table-sign-mantissa] S[round-mode=places,round-precision=2,table-format=4.2] S[round-mode=places,round-precision=2,table-format=1.3,table-sign-mantissa] S[round-mode=places,round-precision=2,table-format=4.2] S[round-mode=places,round-precision=2,table-format=1.3,table-sign-mantissa] S[round-mode=places,round-precision=2,table-format=4.2] S[round-mode=places,round-precision=2,table-format=1.3,table-sign-mantissa]}
\toprule
& &
\multicolumn{4}{c}{\textbf{\Lon\ testing}} &
\multicolumn{4}{c}{\textbf{\Loff\ testing}} \\
\cmidrule(lr){3-6}\cmidrule(lr){7-10}
\textbf{Metric / Species} &
& \multicolumn{2}{c}{No Aux} & \multicolumn{2}{c}{\MaMethode}& \multicolumn{2}{c}{No Aux} & \multicolumn{2}{c}{\MaMethode} \\
\cmidrule(lr){3-4}\cmidrule(lr){5-6}\cmidrule(lr){7-8}\cmidrule(lr){9-10}
& 
& {rRMSE} & {R\textsuperscript{2}} & {rRMSE} & {R\textsuperscript{2}}& {rRMSE} & {R\textsuperscript{2}} & {rRMSE} & {R\textsuperscript{2}} \\
\midrule
\multicolumn{10}{l}{\textbf{Full model}} \\
\midrule
\multirow{4}{*}{\textbf{Volume}} & All & 43.2338 & \bfseries 0.72812 & 43.2707\text{$\downarrow$} & 0.72766\text{$\downarrow$} & 38.6072 & 0.62001 & \bfseries 36.0859\text{$\uparrow$} & 0.66802\text{$\uparrow$} \\
 & Coniferous & 43.6667 & 0.62210 & 43.0494\text{$\uparrow$} & \bfseries 0.63271\text{$\uparrow$} & 39.2850 & 0.57918 & \bfseries 37.5616\text{$\uparrow$} & 0.61529\text{$\uparrow$} \\
 & Deciduous & 41.2399 & \bfseries 0.79781 & 42.3660\text{$\downarrow$} & 0.78661\text{$\downarrow$} & 36.3103 & 0.56104 & \bfseries 26.5117\text{$\uparrow$} & 0.76599\text{$\uparrow$} \\
 & Mixed & 43.3428 & 0.66280 & 42.6917\text{$\uparrow$} & 0.67285\text{$\uparrow$} & \bfseries 34.8927 & \bfseries 0.72404 & 40.5114\text{$\downarrow$} & 0.62801\text{$\downarrow$} \\
\midrule
\multirow{4}{*}{\textbf{Conif. Vol.}} & All & 57.3436 & \bfseries 0.71890 & 59.9568\text{$\downarrow$} & 0.69270\text{$\downarrow$} & \bfseries 44.7815 & 0.66438 & 45.3543\text{$\downarrow$} & 0.65574\text{$\downarrow$} \\
 & Coniferous & 44.7208 & \bfseries 0.61767 & 46.7795\text{$\downarrow$} & 0.58166\text{$\downarrow$} & 40.2900 & 0.57578 & \bfseries 40.2273\text{$\uparrow$} & 0.57710\text{$\uparrow$} \\
 & Deciduous & 347.5516 & -0.52080 & 283.8831\text{$\uparrow$} & \bfseries -0.01464\text{$\uparrow$} & 223.6521 & -0.79010 & \bfseries 215.4588\text{$\uparrow$} & -0.66134\text{$\uparrow$} \\
 & Mixed & 80.7028 & \bfseries 0.40864 & 89.8593\text{$\downarrow$} & 0.26683\text{$\downarrow$} & \bfseries 68.7281 & 0.26799 & 81.9519\text{$\downarrow$} & -0.04079\text{$\downarrow$} \\
\midrule
\multirow{4}{*}{\textbf{Decid. Vol.}} & All & \bfseries 78.7735 & 0.80858 & 81.8622\text{$\downarrow$} & 0.79327\text{$\downarrow$} & 106.0213 & 0.79071 & 96.7532\text{$\uparrow$} & \bfseries 0.82570\text{$\uparrow$} \\
 & Coniferous & \bfseries 534.9853 & \bfseries -1.51052 & 604.7479\text{$\downarrow$} & -2.20795\text{$\downarrow$} & 677.6303 & -1.55400 & 833.6214\text{$\downarrow$} & -2.86521\text{$\downarrow$} \\
 & Deciduous & 42.4786 & \bfseries 0.79209 & 43.7037\text{$\downarrow$} & 0.77993\text{$\downarrow$} & 40.3593 & 0.49434 & \bfseries 28.6177\text{$\uparrow$} & 0.74576\text{$\uparrow$} \\
 & Mixed & 66.1819 & 0.44248 & 66.4016\text{$\downarrow$} & 0.43878\text{$\downarrow$} & \bfseries 63.4189 & \bfseries 0.49107 & 63.9797\text{$\downarrow$} & 0.48203\text{$\downarrow$} \\
\midrule
\multirow{4}{*}{\textbf{Density}} & All & 56.7837 & 0.40255 & \bfseries 56.4560\text{$\uparrow$} & 0.40943\text{$\uparrow$} & 58.1897 & 0.40260 & 56.9036\text{$\uparrow$} & \bfseries 0.42871\text{$\uparrow$} \\
 & Coniferous & 58.0619 & 0.38646 & 57.6329\text{$\uparrow$} & 0.39549\text{$\uparrow$} & 57.7110 & 0.41769 & \bfseries 55.9819\text{$\uparrow$} & \bfseries 0.45206\text{$\uparrow$} \\
 & Deciduous & 52.1140 & 0.42396 & \bfseries 51.6857\text{$\uparrow$} & \bfseries 0.43339\text{$\uparrow$} & 62.0020 & 0.35936 & 61.0460\text{$\uparrow$} & 0.37897\text{$\uparrow$} \\
 & Mixed & \bfseries 52.1315 & 0.30097 & 52.5101\text{$\downarrow$} & 0.29078\text{$\downarrow$} & 56.4164 & \bfseries 0.33184 & 57.9016\text{$\downarrow$} & 0.29620\text{$\downarrow$} \\
\midrule
\multirow{4}{*}{\textbf{Basal Area}} & All & 38.9403 & 0.51094 & 37.7365\text{$\uparrow$} & \bfseries 0.54071\text{$\uparrow$} & 35.5473 & 0.42909 & \bfseries 33.8683\text{$\uparrow$} & 0.48175\text{$\uparrow$} \\
 & Coniferous & 40.6842 & 0.37928 & 38.8367\text{$\uparrow$} & 0.43437\text{$\uparrow$} & 37.0313 & 0.37659 & \bfseries 34.7603\text{$\uparrow$} & \bfseries 0.45071\text{$\uparrow$} \\
 & Deciduous & 35.6143 & 0.65400 & 34.7304\text{$\uparrow$} & \bfseries 0.67096\text{$\uparrow$} & 31.8056 & 0.25482 & \bfseries 26.5063\text{$\uparrow$} & 0.48245\text{$\uparrow$} \\
 & Mixed & 38.8271 & 0.50426 & 39.0266\text{$\downarrow$} & 0.49915\text{$\downarrow$} & \bfseries 29.7533 & \bfseries 0.65071 & 37.9353\text{$\downarrow$} & 0.43220\text{$\downarrow$} \\
\midrule
\multirow{4}{*}{\textbf{Height}} & All & 13.1821 & \bfseries 0.88866 & 13.6922\text{$\downarrow$} & 0.87988\text{$\downarrow$} & 11.8173 & 0.79807 & \bfseries 11.3461\text{$\uparrow$} & 0.81385\text{$\uparrow$} \\
 & Coniferous & 13.6892 & \bfseries 0.87216 & 14.0000\text{$\downarrow$} & 0.86629\text{$\downarrow$} & 11.7594 & 0.78809 & \bfseries 11.5981\text{$\uparrow$} & 0.79386\text{$\uparrow$} \\
 & Deciduous & 12.6147 & \bfseries 0.91514 & 13.6213\text{$\downarrow$} & 0.90106\text{$\downarrow$} & 11.3417 & 0.84785 & \bfseries 9.5543\text{$\uparrow$} & 0.89202\text{$\uparrow$} \\
 & Mixed & 12.2244 & \bfseries 0.89131 & 12.6695\text{$\downarrow$} & 0.88325\text{$\downarrow$} & 12.9022 & 0.79061 & \bfseries 11.4851\text{$\uparrow$} & 0.83408\text{$\uparrow$} \\
\bottomrule
\end{tabular}
\caption{The full model, trained on both \Loff\ and \Lon\ samples, is evaluated on \Lon\ and \Loff\ test sets. Performance is reported in terms of rRMSE and R\textsuperscript{2} for both \MaMethode\ (with auxiliary data) and No Aux configurations. Arrows on \MaMethode\ values indicate whether \MaMethode\ outperforms No Aux ($\uparrow$ \MaMethode\ better, $\downarrow$ No Aux better).}
\end{table}

\section{Results for season-specific model}
\label{app:seasonal_model_results}

\begin{table}[H]
\label{tab:LON_model_full_results}
\centering
\scriptsize
\sisetup{round-mode=places, round-precision=2}
\setlength{\tabcolsep}{4pt}
\begin{tabular}{l l S[round-mode=places,round-precision=2,table-format=4.2] S[round-mode=places,round-precision=2,table-format=1.3,table-sign-mantissa] S[round-mode=places,round-precision=2,table-format=4.2] S[round-mode=places,round-precision=2,table-format=1.3,table-sign-mantissa] S[round-mode=places,round-precision=2,table-format=4.2] S[round-mode=places,round-precision=2,table-format=1.3,table-sign-mantissa] S[round-mode=places,round-precision=2,table-format=4.2] S[round-mode=places,round-precision=2,table-format=1.3,table-sign-mantissa]}
\toprule
& &
\multicolumn{4}{c}{\textbf{Same-season testing}} &
\multicolumn{4}{c}{\textbf{Cross-season testing}} \\
\cmidrule(lr){3-6}\cmidrule(lr){7-10}
\textbf{Metric / Species} &
& \multicolumn{2}{c}{No Aux} & \multicolumn{2}{c}{\MaMethode}& \multicolumn{2}{c}{No Aux} & \multicolumn{2}{c}{\MaMethode} \\
\cmidrule(lr){3-4}\cmidrule(lr){5-6}\cmidrule(lr){7-8}\cmidrule(lr){9-10}
& 
& {rRMSE} & {R\textsuperscript{2}} & {rRMSE} & {R\textsuperscript{2}}& {rRMSE} & {R\textsuperscript{2}} & {rRMSE} & {R\textsuperscript{2}} \\
\midrule
\multicolumn{10}{l}{\textbf{\Lon\ model}} \\
\midrule
\multirow{4}{*}{\textbf{Volume}} & All & 42.0675 & 0.74259 & 41.8049\text{$\uparrow$} & \bfseries 0.74580\text{$\uparrow$} & 39.9638 & 0.59284 & \bfseries 39.4270\text{$\uparrow$} & 0.60370\text{$\uparrow$} \\
 & Coniferous & 40.9931 & \bfseries 0.66696 & 41.6474\text{$\downarrow$} & 0.65625\text{$\downarrow$} & \bfseries 40.4825 & 0.55313 & 40.9747\text{$\downarrow$} & 0.54220\text{$\downarrow$} \\
 & Deciduous & 42.2907 & 0.78737 & 40.3308\text{$\uparrow$} & \bfseries 0.80662\text{$\uparrow$} & 35.8697 & 0.57163 & \bfseries 29.7632\text{$\uparrow$} & 0.70507\text{$\uparrow$} \\
 & Mixed & 41.1572 & \bfseries 0.69595 & 42.3103\text{$\downarrow$} & 0.67867\text{$\downarrow$} & \bfseries 40.7647 & 0.62335 & 43.4931\text{$\downarrow$} & 0.57124\text{$\downarrow$} \\
\midrule
\multirow{4}{*}{\textbf{Conif. Vol.}} & All & 57.1925 & \bfseries 0.72038 & 57.4918\text{$\downarrow$} & 0.71745\text{$\downarrow$} & 51.8923 & 0.54933 & \bfseries 50.3332\text{$\uparrow$} & 0.57600\text{$\uparrow$} \\
 & Coniferous & \bfseries 43.2059 & \bfseries 0.64314 & 43.8139\text{$\downarrow$} & 0.63302\text{$\downarrow$} & 46.9344 & 0.42432 & 45.2995\text{$\uparrow$} & 0.46373\text{$\uparrow$} \\
 & Deciduous & 440.3932 & -1.44183 & 347.8137\text{$\uparrow$} & \bfseries -0.52310\text{$\uparrow$} & 309.6263 & -2.43089 & \bfseries 257.9620\text{$\uparrow$} & -1.38146\text{$\uparrow$} \\
 & Mixed & 84.8848 & \bfseries 0.34576 & 90.2187\text{$\downarrow$} & 0.26096\text{$\downarrow$} & \bfseries 67.6347 & 0.29110 & 76.2298\text{$\downarrow$} & 0.09948\text{$\downarrow$} \\
\midrule
\multirow{4}{*}{\textbf{Decid. Vol.}} & All & \bfseries 82.0085 & \bfseries 0.79253 & 84.1021\text{$\downarrow$} & 0.78180\text{$\downarrow$} & 168.8309 & 0.46927 & 129.4030\text{$\uparrow$} & 0.68822\text{$\uparrow$} \\
 & Coniferous & \bfseries 428.5646 & \bfseries -0.61106 & 601.7601\text{$\downarrow$} & -2.17633\text{$\downarrow$} & 1909.4773 & -19.27986 & 1216.7343\text{$\uparrow$} & -7.23431\text{$\uparrow$} \\
 & Deciduous & 48.0263 & 0.73424 & 45.3316\text{$\uparrow$} & \bfseries 0.76323\text{$\uparrow$} & 43.6752 & 0.40784 & \bfseries 38.0096\text{$\uparrow$} & 0.55150\text{$\uparrow$} \\
 & Mixed & \bfseries 65.5585 & \bfseries 0.45294 & 68.4781\text{$\downarrow$} & 0.40313\text{$\downarrow$} & 72.6716 & 0.33173 & 76.3794\text{$\downarrow$} & 0.26180\text{$\downarrow$} \\
\midrule
\multirow{4}{*}{\textbf{Density}} & All & 55.9299 & 0.42038 & \bfseries 54.1503\text{$\uparrow$} & \bfseries 0.45668\text{$\uparrow$} & 73.6346 & 0.04338 & 63.3664\text{$\uparrow$} & 0.29158\text{$\uparrow$} \\
 & Coniferous & 56.1301 & 0.42660 & \bfseries 54.4680\text{$\uparrow$} & \bfseries 0.46006\text{$\uparrow$} & 76.5391 & -0.02425 & 63.3778\text{$\uparrow$} & 0.29771\text{$\uparrow$} \\
 & Deciduous & 55.6481 & 0.34318 & \bfseries 50.8583\text{$\uparrow$} & \bfseries 0.45138\text{$\uparrow$} & 63.8926 & 0.31970 & 66.2113\text{$\downarrow$} & 0.26942\text{$\downarrow$} \\
 & Mixed & \bfseries 51.7428 & \bfseries 0.31135 & 53.1835\text{$\downarrow$} & 0.27247\text{$\downarrow$} & 62.5444 & 0.17881 & 59.2115\text{$\uparrow$} & 0.26400\text{$\uparrow$} \\
\midrule
\multirow{4}{*}{\textbf{Basal Area}} & All & 38.0134 & 0.53394 & 37.7944\text{$\uparrow$} & \bfseries 0.53930\text{$\uparrow$} & 38.6582 & 0.32479 & \bfseries 37.7709\text{$\uparrow$} & 0.35543\text{$\uparrow$} \\
 & Coniferous & 39.2890 & 0.42112 & \bfseries 38.7924\text{$\uparrow$} & \bfseries 0.43566\text{$\uparrow$} & 40.6156 & 0.25007 & 39.3302\text{$\uparrow$} & 0.29678\text{$\uparrow$} \\
 & Deciduous & 35.9457 & 0.64753 & 34.4637\text{$\uparrow$} & \bfseries 0.67600\text{$\uparrow$} & 29.9156 & 0.34075 & \bfseries 29.7927\text{$\uparrow$} & 0.34616\text{$\uparrow$} \\
 & Mixed & \bfseries 37.2713 & \bfseries 0.54320 & 39.8701\text{$\downarrow$} & 0.47727\text{$\downarrow$} & 37.6851 & 0.43966 & 38.4272\text{$\downarrow$} & 0.41737\text{$\downarrow$} \\
\midrule
\multirow{4}{*}{\textbf{Height}} & All & 13.7651 & \bfseries 0.87860 & 13.9794\text{$\downarrow$} & 0.87479\text{$\downarrow$} & 12.4470 & 0.77598 & \bfseries 12.2183\text{$\uparrow$} & 0.78413\text{$\uparrow$} \\
 & Coniferous & 13.3873 & \bfseries 0.87774 & 14.2411\text{$\downarrow$} & 0.86165\text{$\downarrow$} & 12.8641 & 0.74641 & \bfseries 12.6367\text{$\uparrow$} & 0.75529\text{$\uparrow$} \\
 & Deciduous & 14.7723 & 0.88363 & 14.0027\text{$\uparrow$} & \bfseries 0.89544\text{$\uparrow$} & 11.0423 & 0.85577 & \bfseries 10.2772\text{$\uparrow$} & 0.87507\text{$\uparrow$} \\
 & Mixed & 13.3741 & 0.86990 & 12.9720\text{$\uparrow$} & \bfseries 0.87761\text{$\uparrow$} & \bfseries 10.5525 & 0.85993 & 11.0443\text{$\downarrow$} & 0.84657\text{$\downarrow$} \\
\bottomrule
\end{tabular}
\caption{The \Lon\ model, trained exclusively on \Lon\ samples, is evaluated on a same-season test set (\Lon) and cross-tested on the \Loff\ season. Performance is reported in rRMSE and R\textsuperscript{2} for both \MaMethode\ (with auxiliary data) and No Aux configurations. Arrows on \MaMethode\ values indicate whether \MaMethode\ outperforms No Aux ($\uparrow$ \MaMethode\ better, $\downarrow$ No Aux better).}
\end{table}

\begin{table}[H]
\label{tab:LOFF_model_full_results}
\centering
\scriptsize
\sisetup{round-mode=places, round-precision=2}
\setlength{\tabcolsep}{4pt}
\begin{tabular}{l l S[round-mode=places,round-precision=2,table-format=4.2] S[round-mode=places,round-precision=2,table-format=1.3,table-sign-mantissa] S[round-mode=places,round-precision=2,table-format=4.2] S[round-mode=places,round-precision=2,table-format=1.3,table-sign-mantissa] S[round-mode=places,round-precision=2,table-format=4.2] S[round-mode=places,round-precision=2,table-format=1.3,table-sign-mantissa] S[round-mode=places,round-precision=2,table-format=4.2] S[round-mode=places,round-precision=2,table-format=1.3,table-sign-mantissa]}
\toprule
& &
\multicolumn{4}{c}{\textbf{Same-season testing}} &
\multicolumn{4}{c}{\textbf{Cross-season testing}} \\
\cmidrule(lr){3-6}\cmidrule(lr){7-10}
\textbf{Metric / Species} &
& \multicolumn{2}{c}{No Aux} & \multicolumn{2}{c}{\MaMethode}& \multicolumn{2}{c}{No Aux} & \multicolumn{2}{c}{\MaMethode} \\
\cmidrule(lr){3-4}\cmidrule(lr){5-6}\cmidrule(lr){7-8}\cmidrule(lr){9-10}
& 
& {rRMSE} & {R\textsuperscript{2}} & {rRMSE} & {R\textsuperscript{2}}& {rRMSE} & {R\textsuperscript{2}} & {rRMSE} & {R\textsuperscript{2}} \\
\midrule
\multicolumn{10}{l}{\textbf{\Loff\ model}} \\
\midrule
\multirow{4}{*}{\textbf{Volume}} & All & 39.7127 & 0.59794 & \bfseries 39.6695\text{$\uparrow$} & \bfseries 0.59881\text{$\uparrow$} & 97.7152 & -0.38883 & 65.1479\text{$\uparrow$} & 0.38266\text{$\uparrow$} \\
 & Coniferous & \bfseries 39.4222 & \bfseries 0.57623 & 40.5015\text{$\downarrow$} & 0.55271\text{$\downarrow$} & 130.6374 & -2.38226 & 77.5856\text{$\uparrow$} & -0.19298\text{$\uparrow$} \\
 & Deciduous & 38.8897 & 0.49646 & \bfseries 35.1963\text{$\uparrow$} & 0.58757\text{$\uparrow$} & 48.6952 & 0.71809 & 48.2416\text{$\uparrow$} & \bfseries 0.72332\text{$\uparrow$} \\
 & Mixed & \bfseries 38.9003 & \bfseries 0.65701 & 39.3055\text{$\downarrow$} & 0.64983\text{$\downarrow$} & 78.1879 & -0.09732 & 60.1679\text{$\uparrow$} & 0.35019\text{$\uparrow$} \\
\midrule
\multirow{4}{*}{\textbf{Conif. Vol.}} & All & \bfseries 50.5817 & \bfseries 0.57181 & 54.0259\text{$\downarrow$} & 0.51151\text{$\downarrow$} & 121.9007 & -0.27029 & 85.9591\text{$\uparrow$} & 0.36836\text{$\uparrow$} \\
 & Coniferous & \bfseries 45.3496 & \bfseries 0.46254 & 47.8182\text{$\downarrow$} & 0.40244\text{$\downarrow$} & 97.3651 & -0.81227 & 65.4855\text{$\uparrow$} & 0.18020\text{$\uparrow$} \\
 & Deciduous & \bfseries 298.9264 & -2.19786 & 309.0209\text{$\downarrow$} & -2.41749\text{$\downarrow$} & 645.3675 & -4.24383 & 488.8716\text{$\uparrow$} & \bfseries -2.00901\text{$\uparrow$} \\
 & Mixed & \bfseries 75.7198 & \bfseries 0.11149 & 94.4976\text{$\downarrow$} & -0.38384\text{$\downarrow$} & 155.5240 & -1.19620 & 137.3032\text{$\uparrow$} & -0.71174\text{$\uparrow$} \\
\midrule
\multirow{4}{*}{\textbf{Decid. Vol.}} & All & 150.2513 & 0.57966 & \bfseries 118.7653\text{$\uparrow$} & \bfseries 0.73737\text{$\uparrow$} & 367.2684 & -3.16101 & 196.2938\text{$\uparrow$} & -0.18863\text{$\uparrow$} \\
 & Coniferous & \bfseries 974.5129 & \bfseries -4.28215 & 1009.9941\text{$\downarrow$} & -4.67379\text{$\downarrow$} & 5389.0155 & -253.74067 & 2512.1540\text{$\uparrow$} & -54.35692\text{$\uparrow$} \\
 & Deciduous & 58.6134 & -0.06651 & \bfseries 40.2041\text{$\uparrow$} & 0.49822\text{$\uparrow$} & 73.3575 & 0.37996 & 53.5816\text{$\uparrow$} & \bfseries 0.66920\text{$\uparrow$} \\
 & Mixed & 83.8145 & 0.11109 & \bfseries 65.7736\text{$\uparrow$} & \bfseries 0.45258\text{$\uparrow$} & 215.2816 & -4.89920 & 156.8279\text{$\uparrow$} & -2.13059\text{$\uparrow$} \\
\midrule
\multirow{4}{*}{\textbf{Density}} & All & 59.5450 & 0.37445 & \bfseries 58.6749\text{$\uparrow$} & \bfseries 0.39260\text{$\uparrow$} & 82.1502 & -0.25046 & 67.7876\text{$\uparrow$} & 0.14856\text{$\uparrow$} \\
 & Coniferous & 59.2540 & 0.38613 & \bfseries 57.9761\text{$\uparrow$} & \bfseries 0.41232\text{$\uparrow$} & 83.5360 & -0.27002 & 70.9338\text{$\uparrow$} & 0.08426\text{$\uparrow$} \\
 & Deciduous & 65.5599 & 0.28373 & 63.3144\text{$\uparrow$} & 0.33196\text{$\uparrow$} & 75.1349 & -0.19737 & \bfseries 55.9357\text{$\uparrow$} & \bfseries 0.33637\text{$\uparrow$} \\
 & Mixed & \bfseries 53.0128 & \bfseries 0.41003 & 57.3724\text{$\downarrow$} & 0.30901\text{$\downarrow$} & 78.2819 & -0.57623 & 60.2736\text{$\uparrow$} & 0.06556\text{$\uparrow$} \\
\midrule
\multirow{4}{*}{\textbf{Basal Area}} & All & 36.4631 & 0.39929 & \bfseries 35.7901\text{$\uparrow$} & \bfseries 0.42126\text{$\uparrow$} & 74.7795 & -0.80357 & 53.4985\text{$\uparrow$} & 0.07690\text{$\uparrow$} \\
 & Coniferous & 36.9690 & 0.37868 & \bfseries 36.8813\text{$\uparrow$} & \bfseries 0.38163\text{$\uparrow$} & 92.0718 & -2.17906 & 61.9599\text{$\uparrow$} & -0.43969\text{$\uparrow$} \\
 & Deciduous & 34.9257 & 0.10144 & \bfseries 30.1715\text{$\uparrow$} & 0.32942\text{$\uparrow$} & 42.2649 & 0.51271 & 39.4712\text{$\uparrow$} & \bfseries 0.57500\text{$\uparrow$} \\
 & Mixed & \bfseries 33.8060 & \bfseries 0.54908 & 36.1152\text{$\downarrow$} & 0.48537\text{$\downarrow$} & 61.0204 & -0.22442 & 47.2432\text{$\uparrow$} & 0.26606\text{$\uparrow$} \\
\midrule
\multirow{4}{*}{\textbf{Height}} & All & 16.4798 & 0.60729 & \bfseries 16.0786\text{$\uparrow$} & 0.62618\text{$\uparrow$} & 17.4502 & \bfseries 0.80489 & 18.6028\text{$\downarrow$} & 0.77827\text{$\downarrow$} \\
 & Coniferous & 16.8313 & 0.56588 & \bfseries 16.1537\text{$\uparrow$} & 0.60013\text{$\uparrow$} & 17.4119 & \bfseries 0.79318 & 19.1971\text{$\downarrow$} & 0.74860\text{$\downarrow$} \\
 & Deciduous & \bfseries 14.2663 & 0.75926 & 16.1015\text{$\downarrow$} & 0.69333\text{$\downarrow$} & 18.1251 & \bfseries 0.82482 & 18.1487\text{$\downarrow$} & 0.82436\text{$\downarrow$} \\
 & Mixed & 16.3452 & 0.66394 & \bfseries 15.3839\text{$\uparrow$} & 0.70231\text{$\uparrow$} & 16.4360 & \bfseries 0.80351 & 17.1429\text{$\downarrow$} & 0.78625\text{$\downarrow$} \\
\bottomrule
\end{tabular}
\caption{The \Loff\ model, trained exclusively on \Loff\ samples, is evaluated on a same-season test set (\Loff) and cross-tested on the \Lon\ season. Performance is reported in rRMSE and R\textsuperscript{2} for both \MaMethode\ (with auxiliary data) and No Aux configurations. Arrows on \MaMethode\ values indicate whether \MaMethode\ outperforms No Aux ($\uparrow$ \MaMethode\ better, $\downarrow$ No Aux better).}
\end{table}

\section{Overlapping acquisitions}
\label{app:overlap}
\begin{table}[H]
\footnotesize
\centering
\renewcommand{\arraystretch}{1.2}
\setlength{\tabcolsep}{4pt}
\begin{tabular}{l l cccc}
\toprule
& &
\multicolumn{2}{c}{\textbf{With overlaps}} &
\multicolumn{2}{c}{\textbf{Without overlaps}} \\
\cmidrule(lr){3-4}\cmidrule(lr){5-6}
\textbf{Metric / Species}
& & rRMSE & R\textsuperscript{2}
& rRMSE & R\textsuperscript{2} \\
\midrule
\multirow{4}{*}{\textbf{Volume}}
& All & 52.94 & 0.499 & 58.26 $\pm$ 1.45 & 0.420 $\pm$ 0.029 \\
& Coniferous & 52.01 & 0.412 & 54.02 $\pm$ 3.47 & 0.350 $\pm$ 0.084 \\
& Deciduous & 52.14 & 0.576 & 63.89 $\pm$ 4.98 & 0.431 $\pm$ 0.087 \\
& Mixed & 54.68 & 0.457 & 54.44 $\pm$ 3.56 & 0.439 $\pm$ 0.074 \\
\midrule
\multirow{4}{*}{\textbf{Conif. Vol.}}
& All & 80.87 & 0.324 & 73.15 $\pm$ 1.04 & 0.441 $\pm$ 0.016 \\
& Coniferous & 55.72 & 0.344 & 54.63 $\pm$ 1.41 & 0.364 $\pm$ 0.033 \\
& Deciduous & 1308.53 & -23.456 & 874.49 $\pm$ 147.66 & -11.434 $\pm$ 3.925 \\
& Mixed & 111.06 & -0.164 & 113.14 $\pm$ 2.75 & -0.304 $\pm$ 0.063 \\
\midrule
\multirow{4}{*}{\textbf{Decid. Vol.}}
& All & 135.11 & 0.496 & 147.33 $\pm$ 4.68 & 0.421 $\pm$ 0.037 \\
& Coniferous & 880.14 & -4.212 & 827.85 $\pm$ 297.72 & -4.809 $\pm$ 4.054 \\
& Deciduous & 68.68 & 0.292 & 80.15 $\pm$ 9.12 & 0.130 $\pm$ 0.195 \\
& Mixed & 100.36 & -0.107 & 97.16 $\pm$ 1.55 & -0.200 $\pm$ 0.039 \\
\midrule
\multirow{4}{*}{\textbf{Density}}
& All & 74.58 & 0.033 & 72.88 $\pm$ 1.68 & 0.044 $\pm$ 0.044 \\
& Coniferous & 72.94 & 0.116 & 73.00 $\pm$ 3.06 & 0.085 $\pm$ 0.078 \\
& Deciduous & 79.80 & -0.358 & 76.32 $\pm$ 3.20 & -0.170 $\pm$ 0.097 \\
& Mixed & 74.71 & -0.184 & 65.50 $\pm$ 1.25 & -0.055 $\pm$ 0.040 \\
\midrule
\multirow{4}{*}{\textbf{Basal Area}}
& All & 51.58 & 0.059 & 47.34 $\pm$ 1.19 & 0.212 $\pm$ 0.039 \\
& Coniferous & 52.73 & -0.085 & 47.14 $\pm$ 1.93 & 0.111 $\pm$ 0.073 \\
& Deciduous & 49.77 & 0.195 & 47.64 $\pm$ 5.22 & 0.285 $\pm$ 0.159 \\
& Mixed & 49.45 & 0.140 & 46.03 $\pm$ 0.73 & 0.281 $\pm$ 0.023 \\
\midrule
\multirow{4}{*}{\textbf{Height}}
& All & 19.89 & 0.702 & 20.75 $\pm$ 1.07 & 0.669 $\pm$ 0.035 \\
& Coniferous & 19.58 & 0.688 & 20.25 $\pm$ 0.84 & 0.652 $\pm$ 0.029 \\
& Deciduous & 20.70 & 0.738 & 23.02 $\pm$ 1.56 & 0.690 $\pm$ 0.043 \\
& Mixed & 20.05 & 0.693 & 19.38 $\pm$ 1.52 & 0.701 $\pm$ 0.047 \\
\bottomrule
\end{tabular}
\caption{Effect of spatial overlap between acquisitions on full model performance. Metrics are reported as rRMSE and R\textsuperscript{2} for models trained with auxiliary data (\MaMethode). Results on non-overlapping areas correspond to mean $\pm$ std over $K$ runs.}
\label{tab:overlap-results}
\end{table}

\section{Auxiliary data}
\label{app:Ecological}
\begin{table}[H]
\centering
\scriptsize
\setlength{\tabcolsep}{3pt}
\begin{tabular}{l l c c c c c c c c}
\toprule
\multicolumn{2}{c}{\textbf{Metric / Species}} & \multicolumn{2}{c}{\textbf{No Meta (Ref)}} & \multicolumn{2}{c}{\textbf{Comp.}} & \multicolumn{2}{c}{\textbf{Cover}} & \multicolumn{2}{c}{\textbf{SER}} \\
\cmidrule(lr){1-2}
\cmidrule(lr){3-4}
\cmidrule(lr){5-6}
\cmidrule(lr){7-8}
\cmidrule(lr){9-10}
& & rRMSE & R$^2$ & rRMSE & R$^2$ & rRMSE & R$^2$ & rRMSE & R$^2$ \\
\midrule
\multirow{3}{*}{\textbf{Volume}} & All & 39.78 & 0.7297 & +0.00↓ & +0.0001↑ & -0.34↑ & +0.0046↑ & -0.59↑ & +0.0080↑ \\
 & Coniferous & 40.37 & 0.6384 & -0.52↑ & +0.0093↑ & -0.68↑ & +0.0121↑ & -1.07↑ & +0.0190↑ \\
 & Deciduous & 38.30 & 0.7966 & -0.01↑ & +0.0002↑ & -0.55↑ & +0.0058↑ & -1.16↑ & +0.0122↑ \\
 & Mixed & 38.36 & 0.7226 & +1.68↓ & -0.0249↓ & +1.20↓ & -0.0176↓ & +2.10↓ & -0.0313↓ \\
\midrule
\multirow{3}{*}{\textbf{Conif. Vol.}} & All & 50.73 & 0.7311 & -0.43↑ & +0.0046↑ & -0.87↑ & +0.0092↑ & -1.25↑ & +0.0131↑ \\
 & Coniferous & 41.64 & 0.6307 & -0.58↑ & +0.0102↑ & -0.87↑ & +0.0151↑ & -1.14↑ & +0.0199↑ \\
 & Deciduous & 305.45 & -0.4749 & -41.31↑ & +0.3720↑ & -12.58↑ & +0.1190↑ & -17.78↑ & +0.1667↑ \\
 & Mixed & 73.57 & 0.4490 & +5.31↓ & -0.0825↓ & +1.30↓ & -0.0196↓ & +0.53↓ & -0.0079↓ \\
\midrule
\multirow{3}{*}{\textbf{Decid. Vol.}} & All & 81.86 & 0.8213 & -4.71↑ & +0.0199↑ & -3.33↑ & +0.0142↑ & -5.16↑ & +0.0218↑ \\
 & Coniferous & 540.66 & -1.1938 & -151.64↑ & +1.0580↑ & -22.19↑ & +0.1764↑ & -111.67↑ & +0.8127↑ \\
 & Deciduous & 41.53 & 0.7695 & -1.20↑ & +0.0131↑ & -1.30↑ & +0.0141↑ & -1.83↑ & +0.0199↑ \\
 & Mixed & 61.50 & 0.5193 & -0.05↑ & +0.0008↑ & -3.72↑ & +0.0563↑ & -1.47↑ & +0.0226↑ \\
\midrule
\multirow{3}{*}{\textbf{Density}} & All & 55.15 & 0.4530 & +0.26↓ & -0.0051↓ & +0.27↓ & -0.0054↓ & +0.50↓ & -0.0100↓ \\
 & Coniferous & 56.38 & 0.4553 & -0.29↑ & +0.0056↑ & -0.65↑ & +0.0123↑ & +0.87↓ & -0.0169↓ \\
 & Deciduous & 51.69 & 0.4644 & +1.74↓ & -0.0366↓ & +2.09↓ & -0.0441↓ & -0.70↑ & +0.0145↑ \\
 & Mixed & 51.17 & 0.3565 & +1.63↓ & -0.0415↓ & +3.27↓ & -0.0848↓ & -0.21↑ & +0.0053↑ \\
\midrule
\multirow{3}{*}{\textbf{Basal Area}} & All & 36.76 & 0.5255 & -0.22↑ & +0.0057↑ & -0.37↑ & +0.0095↑ & -0.87↑ & +0.0223↑ \\
 & Coniferous & 38.01 & 0.4228 & -0.50↑ & +0.0153↑ & -0.38↑ & +0.0116↑ & -0.92↑ & +0.0278↑ \\
 & Deciduous & 33.99 & 0.6402 & -0.25↑ & +0.0052↑ & -0.83↑ & +0.0173↑ & -1.79↑ & +0.0369↑ \\
 & Mixed & 36.09 & 0.5579 & +0.87↓ & -0.0216↓ & +0.42↓ & -0.0104↓ & +0.79↓ & -0.0198↓ \\
\midrule
\multirow{3}{*}{\textbf{Height}} & All & 12.16 & 0.8869 & -0.41↑ & +0.0074↑ & +0.04↓ & -0.0009↓ & -0.47↑ & +0.0084↑ \\
 & Coniferous & 12.30 & 0.8718 & -0.20↑ & +0.0043↑ & +0.24↓ & -0.0049↓ & -0.35↑ & +0.0074↑ \\
 & Deciduous & 11.96 & 0.9166 & -1.11↑ & +0.0147↑ & -0.44↑ & +0.0061↑ & -1.04↑ & +0.0138↑ \\
 & Mixed & 11.73 & 0.8912 & -0.36↑ & +0.0064↑ & -0.16↑ & +0.0029↑ & -0.13↑ & +0.0023↑ \\
\midrule
\bottomrule
\end{tabular}
\caption{Ecological auxiliary variables (composition, cover, SER).}
\end{table}

\label{app:SpatioTemporal}

\begin{table}[H]
\centering
\scriptsize
\setlength{\tabcolsep}{3pt}
\begin{tabular}{l l c c c c c c c c}
\toprule
\multicolumn{2}{c}{\textbf{Metric / Species}} & \multicolumn{2}{c}{\textbf{No Meta (Ref)}} & \multicolumn{2}{c}{\textbf{Center}} & \multicolumn{2}{c}{\textbf{Date}} & \multicolumn{2}{c}{\textbf{Delta}} \\
\cmidrule(lr){1-2}
\cmidrule(lr){3-4}
\cmidrule(lr){5-6}
\cmidrule(lr){7-8}
\cmidrule(lr){9-10}
& & rRMSE & R$^2$ & rRMSE & R$^2$ & rRMSE & R$^2$ & rRMSE & R$^2$ \\
\midrule
\multirow{4}{*}{\textbf{Volume}} & All & 39.78 & 0.7297 & -1.45↑ & +0.0194↑ & +0.45↓ & -0.0061↓ & -0.09↑ & +0.0012↑ \\
 & Coniferous & 40.37 & 0.6384 & -1.20↑ & +0.0212↑ & -0.39↑ & +0.0070↑ & -0.27↑ & +0.0048↑ \\
 & Deciduous & 38.30 & 0.7966 & -3.34↑ & +0.0339↑ & +0.48↓ & -0.0051↓ & -0.31↑ & +0.0033↑ \\
 & Mixed & 38.36 & 0.7226 & +1.42↓ & -0.0210↓ & +3.02↓ & -0.0454↓ & +0.94↓ & -0.0138↓ \\
\midrule
\multirow{4}{*}{\textbf{Conif. Vol.}} & All & 50.73 & 0.7311 & -1.99↑ & +0.0207↑ & -0.30↑ & +0.0032↑ & -0.77↑ & +0.0081↑ \\
 & Coniferous & 41.64 & 0.6307 & -1.96↑ & +0.0340↑ & -0.81↑ & +0.0141↑ & -0.37↑ & +0.0064↑ \\
 & Deciduous & 305.45 & -0.4749 & +4.95↓ & -0.0482↓ & +30.59↓ & -0.3102↓ & -35.79↑ & +0.3254↑ \\
 & Mixed & 73.57 & 0.4490 & -0.17↑ & +0.0026↑ & +3.67↓ & -0.0564↓ & -2.03↑ & +0.0299↑ \\
\midrule
\multirow{4}{*}{\textbf{Decid. Vol.}} & All & 81.86 & 0.8213 & -5.49↑ & +0.0231↑ & -0.20↑ & +0.0009↑ & -5.73↑ & +0.0241↑ \\
 & Coniferous & 540.66 & -1.1938 & -40.13↑ & +0.3136↑ & -151.45↑ & +1.0569↑ & -99.42↑ & +0.7327↑ \\
 & Deciduous & 41.53 & 0.7695 & -4.25↑ & +0.0447↑ & +1.97↓ & -0.0225↓ & -1.40↑ & +0.0152↑ \\
 & Mixed & 61.50 & 0.5193 & +0.38↓ & -0.0061↓ & +2.11↓ & -0.0336↓ & -5.10↑ & +0.0764↑ \\
\midrule
\multirow{4}{*}{\textbf{Density}} & All & 55.15 & 0.4530 & +1.35↓ & -0.0271↓ & -0.16↑ & +0.0031↑ & +0.55↓ & -0.0111↓ \\
 & Coniferous & 56.38 & 0.4553 & +0.97↓ & -0.0191↓ & -0.37↑ & +0.0070↑ & +0.16↓ & -0.0031↓ \\
 & Deciduous & 51.69 & 0.4644 & +1.41↓ & -0.0296↓ & +0.67↓ & -0.0139↓ & +0.47↓ & -0.0097↓ \\
 & Mixed & 51.17 & 0.3565 & +3.41↓ & -0.0885↓ & +0.02↓ & -0.0005↓ & +2.94↓ & -0.0760↓ \\
\midrule
\multirow{4}{*}{\textbf{Basal Area}} & All & 36.76 & 0.5255 & -1.22↑ & +0.0310↑ & +0.19↓ & -0.0050↓ & -0.06↑ & +0.0014↑ \\
 & Coniferous & 38.01 & 0.4228 & -1.11↑ & +0.0334↑ & -0.32↑ & +0.0098↑ & -0.14↑ & +0.0044↑ \\
 & Deciduous & 33.99 & 0.6402 & -2.55↑ & +0.0519↑ & +0.72↓ & -0.0154↓ & -0.50↑ & +0.0105↑ \\
 & Mixed & 36.09 & 0.5579 & +0.51↓ & -0.0126↓ & +1.24↓ & -0.0311↓ & +0.99↓ & -0.0248↓ \\
\midrule
\multirow{4}{*}{\textbf{Height}} & All & 12.16 & 0.8869 & -0.08↑ & +0.0013↑ & -0.23↑ & +0.0041↑ & -0.32↑ & +0.0058↑ \\
 & Coniferous & 12.30 & 0.8718 & +0.08↓ & -0.0017↓ & -0.04↑ & +0.0010↑ & -0.19↑ & +0.0040↑ \\
 & Deciduous & 11.96 & 0.9166 & -0.59↑ & +0.0079↑ & -0.71↑ & +0.0096↑ & -0.87↑ & +0.0117↑ \\
 & Mixed & 11.73 & 0.8912 & -0.07↑ & +0.0012↑ & -0.37↑ & +0.0067↑ & -0.12↑ & +0.0021↑ \\
\midrule
\bottomrule
\end{tabular}
\caption{Spatiotemporal auxiliary variables (center, date, delta effects).}
\end{table}



\bibliographystyle{elsarticle-harv} 
\bibliography{references}



\end{document}